\documentclass[final,12pt]{msml2022} % Anonymized submission
\title[Decision Boundaries and Convex Hulls in the Feature Space]{Decision Boundaries and Convex Hulls in the Feature Space\\ that Deep Learning Functions Learn from Images}
\usepackage{times}
\msmlauthor{%
 \Name{Roozbeh Yousefzadeh} \Email{roozbeh.yousefzadeh@yale.edu}\\
 \addr Yale Center for Medical Informatics and VA CT Healthcare System
}

\DeclareMathOperator{\softmax}{softmax}

\newcommand\hull{$\mathcal{H}^{tr}$\xspace}

\usepackage{xspace,xcolor,bm,amsmath,graphicx,bbm,booktabs}

\makeatletter
 \let\Ginclude@graphics\@org@Ginclude@graphics
\makeatother

\begin{document}

\vspace{-1cm}

\maketitle

\vspace{-.8cm}

\begin{abstract}%
The success of deep neural networks in image classification and learning can be partly attributed to the features they extract from images. It is often speculated about the properties of a low-dimensional manifold that models extract and learn from images. However, there is not sufficient understanding about this low-dimensional space based on theory or empirical evidence. For image classification models, their last hidden layer is the one where images of each class is separated from other classes and it also has the least number of features. Here, we develop methods and formulations to study that feature space for any model. We study the partitioning of the domain in feature space, identify regions guaranteed to have certain classifications, and investigate its implications for the pixel space. We observe that geometric arrangements of decision boundaries in feature space is significantly different compared to pixel space, providing insights about adversarial vulnerabilities, image morphing, extrapolation, ambiguity in classification, and the mathematical understanding of image classification models.
\end{abstract}


\vspace{-.4cm}

\section{Introduction}

\vspace{-.2cm}

The process in which deep networks learn to classify images is not adequately understood. In the context of classification, successful learning can be described as learning the similarities and differences between samples of each class. But for images, similarities and differences usually cannot be identified or explained in terms of individual pixels. So, how do models and humans identify similarities and see differences in images? The spatial relationship between groups of pixels and the patterns that are depicted via such pixel groups are instrumental in classifying them by humans and models. If we ask a person why they classify a particular image as a cat, they might point out the specific patterns such as the shape of ears and eyes of the cat. If we ask a radiologist why they classify a tumor as cancerous, they might point out the shape of the tumor and the patterns visible in that region. Analyzing these patterns can be considered feature extraction, and those features, as opposed to individual pixels, would be the ones helpful for learning and classification. 

In deep learning, feature extraction is performed via specialized computational tools, i.e., convolutional layers, and it is not easy to disentangle the feature extraction from the learning process as a whole. Often, when a model has good generalization accuracy, one considers that the model has learned some useful features \citep{chen2021feature}, but it is not clear what those features are \citep{berner2021modern}. This lack of understanding is evident when we consider vulnerability of models to adversarial examples \citep{shafahi2018adversarial}. Sometimes adversarial examples are themselves considered features \citep{ilyas2019adversarial}. Another issue arises when one gives out-of-distribution images to a model, e.g., a model trained for object recognition may classify a radiology image of liver as Airplane with 100\% confidence, defying the notion of learning. Despite these shortcomings, deep networks are impressively successful in a wide range of tasks related to image classification, e.g., facial recognition, object recognition, medical imaging. There have been several studies to improve our understanding of what models learn from images, e.g., \citet{xiao2020noise} studied the effect of image backgrounds. Several other studies focused on verifying whether models have learned generalizable features \citep{yadav2019cold,recht2018cifar,recht2019imagenet}. \citet{neyshabur2020towards} used feedforward networks to learn the convolutional filters from scratch. \citet{alain2016understanding} studied linear separability of the classes in intermediate layers of trained networks. \citet{balestriero2018mad} showed that deep neural networks are spline operators that partition their domain. \citet{recanatesi2021predictive} studied feedforward networks and concluded that models learn a low-dimensional latent representation from images. This idea is pursued before under a field known as representation learning \citep{bengio2013representation,oord2018representation}. There are studies on geometry of data and the separability of classes, e.g., \citep{mallat2016understanding,cohen2020separability,fawzi2018empirical,bronstein2017geometric}. Moreover, specific deep learning architectures are introduced that process images with wavelet scattering \citep{bruna2013invariant,zarka2019deep} to provide a way to understand properties of the features learned by the models, e.g., \cite{zarka2021separation} studied the Fisher discriminant ratio of learned features.% These scattering architectures provide useful insights, however, geometric arrangements of decision boundaries of image classification models are not studied in the context of those models.

% \cite{bruna2013invariant} proposed wavelet scattering networks which first extracts features from images then proceeds with classification. Such approach has been used , but scattering networks do not compete with modern architectures for image classification.

% \cite{bubba2019learning} shearlets

% \cite{} considered the possible advantages of using wavelets to learn the convolutional filters that extract features from images. 

In this paper, we develop methods to complement the previous work and provide a better understanding of the feature space that deep networks learn from images. We consider the last hidden layer of image classification models as the feature space with least number of features where images of each class are separated from other classes. Our contributions can be summarized as:
    \vspace{-.2cm}
\begin{enumerate}
    \item We develop methods and formulations that can be used to systematically investigate the feature space learned by any trained model. We investigate how images map to the feature space, and how that feature space relates to the pixel space. Finding images in the pixel space that would directly map to particular points and regions in the feature space is an inverse problem involving the trained models, the type of problem that is generally considered hard to solve \citep{elsayed2018large}. We use the homotopy algorithm by \citet{yousefzadeh2020deep} to solve our formulations.
    \vspace{-.2cm}
 %   \item We develop methods to study decision boundaries of models in feature space and verify the robustness of models in that space. We see that decision boundaries are much farther away from the samples, as opposed to the pixel space. As a result, models are less vulnerable to adversarial examples in the feature space. Our method can identify a ball around each sample where the classification of model is guaranteed to be constant. This leads us to investigate unions of regions in the feature space with respect to partitions for each class.
    \item We study the functional task of models in that feature space and see that testing samples are all outside the convex hull of training set even in a 64-dimensional feature space learned by the models, i.e., functional task of models involve moderate extrapolation. We previously reported that image classification requires extrapolation both in pixel space and in the feature space \citep{yousefzadeh2021hull}. More recently, \citet{balestriero2021learning} concluded that in high-dimensional space (larger than 100 dimensions), learning always amounts to extrapolation. Our results in this paper show that even in a 64-dimensional space learned by the models, image classification still requires extrapolation.
    \vspace{-.2cm}
    \item Our method identifies points in the pixel space that would map to decision boundaries and convex hulls in the feature space providing novel insights about the functional performance of models in that space, and the extent of extrapolation. We observe that arrangements of decision boundaries and convex hulls in feature space differ from the pixel space in meaningful ways, not reported in the literature. Our methods can also be used for image morphing.
    \vspace{-.2cm}
    \item We propose a new method to identify ambiguous and adversarial images based on their relative distance to decision boundaries and the convex hull of training set in the feature space. In the feature space, unlike the pixel space, most testing images are relatively close to the convex hull of training set while far from the decision boundaries. Ambiguous images, however, are close to decision boundaries and far from the convex hull. Adversarial inputs are also recognizably close the decision boundaries of feature space. Moreover, adversarial methods such as DeepFool \citep{moosavi2016deepfool} and PGD attacks \citep{madry2018towards} move images towards the convex hull of training set in the feature space.% Our empirical results show that testing samples are always outside that convex hull while adversarial examples are either close to or inside it.% Adversarial examples designed base on knowledge of training set also fall inside that hull. As a result, an input falling inside the convex hull of training
\end{enumerate}

\vspace{-.6cm}

\section{Feature space learned by trained models} \label{sec:featurespace}

\vspace{-.2cm}

We consider the feature space in the last hidden layer of trained models. This feature space is the key to successful classification of images and it usually has the least dimensionality compared to other hidden layers. Our trained model is a function denoted by $\mathcal{N}(.)$ that operates on input images and produces an output vector
\begin{equation} \label{eq:map2z}
    z = \mathcal{N}(x),
\end{equation}
where each element of $z$ corresponds to one class, and the class(es) with the largest value will be the classification of the model\footnote{For brevity, we may sometimes use $\mathcal{C}(x)$ to denote the classification of the model for $x$, implying that a $z$ has been computed for $x$ and $\mathcal{C}(.)$ has been applied to that $z$.}
\begin{equation}
    \mathcal{C}(z) = \{ i : z_i = \max_k z_k \}.
\end{equation}
Domain of $\mathcal{N}$ is denoted by $\Omega$ which would be the pixel space for image classification models. Any given model is trained to recognize a certain number of classes. In our notation, pixel space has $p$ dimensions/pixels and $z$ has $n$ elements/classes.

We use $\Phi$ to denote the feature space in the last hidden layer of $\mathcal{N}$. An input image, $x$, has a mapping to that feature space denoted by $x_\phi$. We can formalize this mapping via our trained model
% \vspace{-.2cm}
\begin{equation} \label{eq:map2f}
	x_\phi = \mathcal{N}_\phi (x),
\end{equation}
where $\Phi$ has $f$ dimensions. $\mathcal{N}_\phi (.)$ is similar to $\mathcal{N} (.)$ except that it returns the output of the last hidden layer of the model. Similar to pixel space, feature space will also have a domain, $\Omega_\phi$ which would be the range of $\mathcal{N}_\phi (.)$.

After the last hidden layer, the model has a fully connected layer and a softmax layer. Hence, the output of the model, $z$, can be written in terms of the feature space:
\begin{equation} \label{eq:soft}
    z = \softmax(x_\phi W_\phi + b_\phi),
\end{equation}
where $W_\phi$ is the weight matrix for the last fully connected layer, with $f$ rows and $n$ columns, and $b_\phi$ is the bias vector for that layer with $n$ elements. It is sensible to assume $n < f$, i.e., feature space has more dimensions than the number of output classes.% Again, the element of $z$ with largest value would be the classification of the model.%, and for brevity we may use $\mathcal{C}(x_\phi)$ to denote the classification of the model for $x_\phi$.

%whether $z$ is obtained directly via equation~\eqref{eq:map2z}, or indirectly via equation~\eqref{eq:soft}. Hence, any $x_\phi \in \Omega_\phi$ will also have a classification via $\mathcal{C}(x_\phi) =$.

Our following formulations are applicable to any model with any number of features in its hidden layers, i.e., $\mathcal{N}$ can be any trained model. Moreover, one can study the feature space in any of the hidden layers, though, in this work, our focus is on the last hidden layer. To make this more tangible, consider $\mathcal{N}$ to be a standard CNN, pre-trained on CIFAR-10 dataset. Model has a standard residual network architecture \citep{he2016deep} with total depth of 20 layers while the last hidden layer has 64 features.\footnote{\scriptsize Pre-trained model is available at \url{https://www.mathworks.com/help/deeplearning/ug/train-residual-network-for-image-classification.html}.\label{fn:model}} It follows that $\Phi$ for this particular model has 64 dimensions. We choose this model because its last hidden layer has fewer features than the standard ResNet-18.% Though in our experiments we only consider the last hidden layer of $\mathcal{N}$, in principle, $\Phi$ can be the feature space in any of the hidden layers.

For a given $x$, one can easily compute its corresponding $x_\phi$ (i.e., map $x$ to $\Phi$) by feeding $x$ to the trained model and computing the output of the model's last hidden layer. However, given an arbitrary $x_\phi$, it is not as easy to find its corresponding $x$ in the pixel space. That is, a trained model $\mathcal{N}$, and by extension $\mathcal{N}_\phi$, are not invertible, i.e., there is not an inverse function $\mathcal{N}_\phi^{-1}$ readily available to map an arbitrary $x_\phi$ to the pixel space. Moreover, the mapping from the pixel space to $\Phi$ is not one-to-one.\footnote{This can be easily verified via any of the pooling layers.}

In Sections~\ref{sec:map2ball}-\ref{sec:map2point}, we will formulate and solve optimization problems to find images (in the pixel space) that would directly map to particular points and regions in the feature space. Before that, let us formulate the decision boundaries of the model in the feature space.

% \vspace{-.2cm}

\section{Decision boundaries in the feature space} \label{sec:boundaries_fs}

An image classification model is a classification function that partitions its domain and assigns a class to each partition \citep{strang2019linear}. Partitions are defined by decision boundaries and so is the model. We can study the decision boundaries and partitions of the model, not just in the pixel domain, but also in the feature space $\Phi$. A point on the decision boundary between classes $i$ and $j$ would be a point that satisfies
% \vspace{-.1cm}
\begin{equation} \label{eq:boundary1}
    z_i = z_j,
\end{equation}
% \vspace{-.4cm}
\begin{equation} \label{eq:boundary2}
    z_i \geq z_k, \forall k \notin \{ i,j \}.
\end{equation}
Any point that satisfies the conditions above will be a flip between classes $i$ and $j$, so we call it a flip point \citep{yousefzadeh2020deep,yousefzadeh2020auditing}. We denote flip points by $x^{f(i,j)}$ when they are in the pixel space, and denote them by $x^{f(i,j)}_\phi$ when they are in the feature space. 

For the purpose of identifying points on the decision boundaries of the model, we can ignore the softmax operation in equation~\eqref{eq:soft} because it only normalizes the values of $z$ to be between 0 and 1, and does not change their order. Therefore, in the following, we will drop the softmax from equation~\eqref{eq:soft} because it does not have an effect on satisfying constraints~\eqref{eq:boundary1}-\eqref{eq:boundary2}. As a result $x_\phi^{f(i,j)}$ should satisfy
% \vspace{-.1cm}
\begin{equation} \label{eq:fb_c1}
    x_\phi^{f(i,j)} W_\phi(:,i) + b_\phi(i) = x_\phi^{f(i,j)} W_\phi(:,j) + b_\phi(j),
\end{equation}
% \vspace{-0.8cm}
\begin{equation} \label{eq:fb_c2}
    x_\phi^{f(i,j)} W_\phi(:,i) + b_\phi(i) \geq x_\phi^{f(i,j)} W_\phi(:,k) + b_\phi(k), \; \forall k \notin \{ i,j \},
\end{equation}
where $W_\phi(:,i)$ denotes the $i^{th}$ column of $W_\phi$.

For a given model, there usually are infinite number of $x_\phi^{f(i,j)}$ satisfying the constraints~\eqref{eq:fb_c1}-\eqref{eq:fb_c2}, but we may be interested to find the $x_\phi^{f(i,j)}$ that is closest to a particular $x_\phi$. Consider that element $i$ of $z$ has the largest value for the input $x$, i.e., classification of $x$ and $x_\phi$ are $i$. The \textit{closest} flip point to $x_\phi$ between classes $i$ and $j$ is denoted by $x_\phi^{f(i,j),c}$ and obtained via the objective function
\begin{equation} \label{eq:fb_obj}
    \min_{x_\phi^{f(i,j),c}} \| x_\phi - x_\phi^{f(i,j),c} \|^2_2.
\end{equation}
Our feature space $\Phi$ is usually lower bounded by zero because it is the result of convolutional, ReLU, and max pooling layers. Hence, we require
\begin{equation} \label{eq:fb_c3}
    0 \leq x_\phi^{f(i,j),c}.
\end{equation}
The optimization problem defined by objective function~\eqref{eq:fb_obj} subject to constraints \eqref{eq:fb_c1},\eqref{eq:fb_c2}, and \eqref{eq:fb_c3} is convex, and there are reliable algorithms to solve it. In most cases, it may be strictly convex, making the optimal solution unique. Either way, the minimum distance to decision boundaries (a.k.a. margin) will be a unique value. The minimum distance of $x_\phi$ to the decision boundary between classes $i$ and $j$ is
\begin{equation}
    d^{f(i,j)}_\phi(x_\phi) = \| x_\phi - x_\phi^{f(i,j),c} \|_2.
\end{equation}
For a model with $n$ output classes and for a specific input $x$, mapped to $x_\phi$ and classified as $i$, we can compute its margin in $\Phi$ to all other $n-1$ classes and find out decision boundary of which class is closest to it. We denote the closest margin by
\begin{equation} \label{eq:fb_dmin}
    d^{f,min}_\phi (x_\phi) = \min_{j \in \{1:n \backslash i\} } d^{f(i,j)}_\phi(x_\phi).
\end{equation}
% which corresponds to the closes flip point in feature space.

% \vspace{-.5cm}

\begin{figure}[h]
    \centering
    \includegraphics[width=0.38\linewidth]{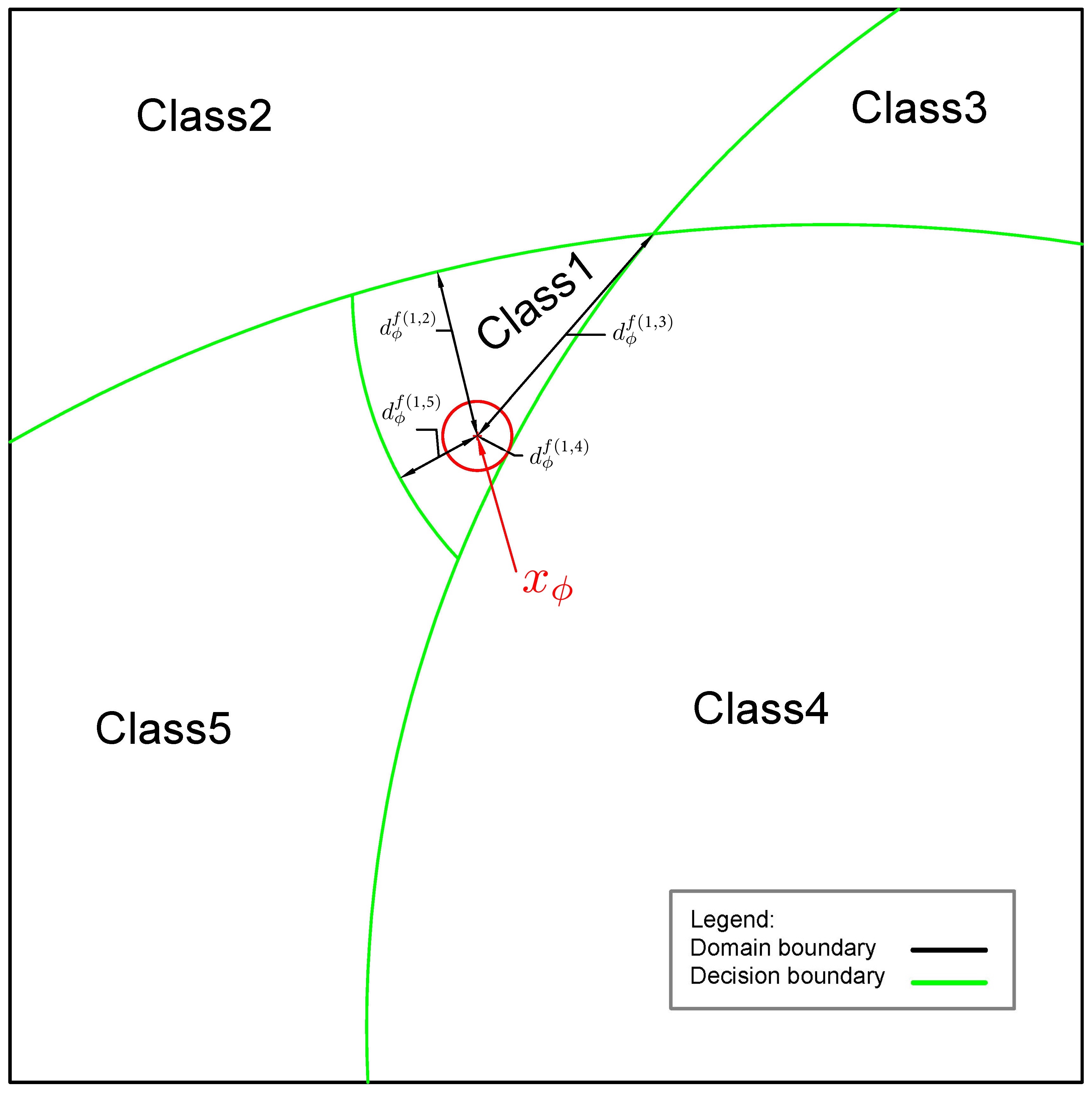}
    \vspace{-.4cm} 
    \caption{An example 2D domain with 5 partitions. Input $x_\phi$, in the feature space, is located in the partition for Class 1. Its margin to each of the other classes is marked. The red ball is the largest ball centered around $x_\phi$ where every point inside it is guaranteed to have the same classification as $x_\phi$.}
    \label{fig:fs_2Dsketch}
\end{figure}

Consider, for example, the 2D domain depicted in Figure~\ref{fig:fs_2Dsketch} which has 5 partitions representing 5 different classes. Input $x$ is located in the partition associated with Class 1. This particular input has a margin to each of the other four classes and the minimum margin is to Class 4. Since our optimization problems in $\Phi$ are convex, we can calculate $d^{f,min}_\phi$ precisely and be sure that it actually is the distance to the closest decision boundary.
%\begin{figure}[h]
%  \centering
%  \label{fig:fs_2Dsketch}
%   \includegraphics[width=0.38\linewidth]{figures/2D_domain_boundaries.jpg}
%   \caption{An example 2D domain with 5 partitions. Input $x_\phi$, in the feature space, is located in the partition for Class 1. Its margin to each of the other classes is marked. The red ball is the largest ball centered around $x_\phi$ where every point inside it is guaranteed to have the same classification as $x_\phi$.}
%\end{figure}

%Figure~\ref{fig:my_label}

Let us now consider the ball centered at $x_\phi$ with radius $d^{f,min}_\phi$, and denote it by $\mathcal{B}(x_\phi)$. Such ball may be entirely inside the domain of feature space, $\Omega_\phi$, or it may extend outside the domain, if $x_\phi$ is close to the boundaries of the $\Omega_\phi$ in some dimensions. Either way, classification of $\mathcal{N}$ for the entire region inside the intersection of $\mathcal{B}(x_\phi)$ and $\Omega_\phi$ is guaranteed to be the same as the classification for $x$ and $x_\phi$
\begin{equation} \label{eq:fb_ball}
    \forall y_\phi \in (\mathcal{B}(x_\phi) \cap \Omega_\phi): \mathcal{C}(y_\phi) = \mathcal{C}(x_\phi),
\end{equation}
i.e., any point in $\Omega_\phi$ that its distance to $x_\phi$ is less than $d^{f,min}_\phi (x_\phi)$ has the same classification as $x_\phi$. 
For this guarantee, we note that $\Phi$ is a continuous space and the output of $\mathcal{N}$ is Lipschitz continuous with respect to points in $\Phi$. In fact, Lipschitz constant of the model with respect to $\Phi$ would be $\sigma_{max}(W_\phi)$, i.e., the largest singular value of $W_\phi$, since one can prove
\begin{equation}
    \| z(x_\phi) - z(y_\phi) \|_2 \leq \sigma_{max}(W_\phi) \| x_\phi - y_\phi \|_2,
\end{equation}
for any $x_\phi$ and $y_\phi$ in feature space.

The radius of the ball $\mathcal{B}(x_\phi)$ gives a measure of robustness for the classification of the model with respect to perturbations in feature space. By studying the decision boundaries, one can also design and analyze adversarial inputs in the feature space and then trace them back to the pixel space as we will explore in numerical experiments. In the following two sections, we provide formulations to reveal the relationship between feature space and pixel space.

% write kkt conditions
% unique optimal solution
% theory some equations

% \subsection{Margin to decision boundaries of feature space}

% Margin for a given input is its distance to the closest decision boundary \citep{elsayed2018large}. 

% smallest margin gives the radius of a ball where all points inside the ball belong to the given class

% any image in the pixel space that maps to the ball will still be

% Next, we formulate and solve optimization problems involving $N_\phi$ to gain an understanding of the feature space $\Phi$.

\section{Seeking points in the pixel space that would map to particular regions in the feature space} \label{sec:map2ball}

As the first goal, let us find images in the pixel space that would map to particular regions in the feature space. Specifically, we seek to find images in the pixel space that would map to $\mathcal{B}(x_\phi) \cap \Omega_\phi$ around any particular image $x_\phi$ satisfying equation~\eqref{eq:fb_ball}. The following constraints will ensure such mapping for $x^{\Omega \rightarrow \mathcal{B}(x_\phi)}$
\begin{equation} \label{eq:ball_c1}
    \| \mathcal{N}_\phi(x^{\Omega \rightarrow \mathcal{B}(x_\phi)}) - x_\phi \|_2 < \lambda \quad , \quad x \in \Omega,
\end{equation}
where $\lambda$ is the radius of the $\mathcal{B}$ or region of interest.

Many different images (in pixel space) may satisfy the constraint above for a particular $x_\phi$ as we shall see in experimental results. To gain an understanding of the variety of such images, we seek to find the ones that are closest to a reference point, $x^r$, in pixel space. A reference point may be any training or testing image, or any other image such as a completely black or white image. Minimizing the distance to reference point is our objective function
\begin{equation} \label{eq:ball_obj}
    \min_{x^{\Omega \rightarrow \mathcal{B}(x_\phi)}} \| x^{\Omega \rightarrow \mathcal{B}(x_\phi)} - x^r \|^2_2,
\end{equation}
and our constraint is equation~\eqref{eq:ball_c1}. We can solve this optimization problem for various reference points, $x^r$, to gain an understanding of the ball surrounding the sample $x_\phi$.

Unlike our set of optimization problems in Section~\ref{sec:boundaries_fs}, optimization problems in Sections \ref{sec:map2ball} and \ref{sec:map2point} may be non-convex because they involve a typically non-convex function $\mathcal{N}_\phi$. Hence, it is important that global optimization algorithms be utilized for solving them. Moreover, issue of vanishing and exploding gradients \citep{bengio1994learning} may arise which is addressed in our previous work \citep{yousefzadeh2020deep}.% A homotopy algorithm is also provided to overcome those issues when solving optimization problems incorporating a deep learning function $\mathcal{N}$ \citep{yousefzadeh2020deep}.

% In previous work, we have developed a homotopy algorithm for solving optimization problems involving deep neural networks, and made recommendations about the use of other algorithms \citep{yousefzadeh2020deep,yousefzadeh2020auditing}.

\section{Seeking points in the pixel space that would map to particular points in feature space} \label{sec:map2point}

We now seek points in the pixel space that $\mathcal{N}_\phi$ will directly map them to a particular $x_\phi$. For an input $x^{\Omega \rightarrow x_\phi}$, this condition can be formalized as:
\begin{equation} \label{eq:point_c1}
 \mathcal{N}_\phi(x^{\Omega \rightarrow x_\phi}) = x_{\phi} \quad , \quad x^{\Omega \rightarrow x_\phi} \in \Omega.
\end{equation}
The particular $x_\phi$ may be any point of interest in the feature space, for example, a point on a decision boundary, or a point on the boundary of a convex hull.

It is possible that $x^{\Omega \rightarrow x_\phi}$ defined by equations~\eqref{eq:point_c1} is not unique, rather, a region, $\mathcal{S}^{\Omega \rightarrow x_{\phi}}$, in the pixel space (contiguous or not), will all map to a particular point in the feature space. We seek to find the $x^{h,\phi}_{\Omega}$ that is {\em closest} to a reference point $x^r$ using the objective function
\begin{equation} \label{eq:point_obj}
  \min_{x^{\Omega \rightarrow x_\phi}} \|  x^{\Omega \rightarrow x_\phi} - x^r \|,
\end{equation}
subject to constraint~\eqref{eq:point_c1}.

% To gain a better understanding of $\mathcal{S}_{\Omega \leftrightarrow x^h_{\phi}}$, we can seek to find other points in that region, for example, the point {\em farthest} from $x$ that would map to the same $x^h_{\phi}$. For finding such point, we modify our objective function in \eqref{eq:f_obj} to

% \begin{equation} \label{eq:point_obj_max}
%   \max_{x^{h,\phi,f}_{\Omega}} \|  x - x^{h,\phi,f}_{\Omega} \|,
% \end{equation}
% and solve it with the same constraints~\eqref{eq:point_c1}. The solution to this maximization problem will be the point in $\Omega$ farthest from $x$ that maps directly to $x^h_\phi$.

It is sensible to use a reference point that has the same classification as $x_\phi$. In such case, we can impose an additional constraint to ensure $x^{\Omega \rightarrow x_\phi}$ and $x^r$ belong to the same partition in pixel space.
\begin{equation} \label{eq:f_c3}
  \exists \pi , \pi : (x^{\Omega \rightarrow x_\phi} , x^r) \: | \: \forall x \in \pi,  \mathcal{C}(\mathcal{N}(x)) = \mathcal{C}(\mathcal{N}(x^r)),
\end{equation}
To verify the additional constraint \eqref{eq:f_c3}, one needs to verify Lipschitz continuity of $\mathcal{N}$ not just in the feature but also in the pixel space $\Omega$. There are methods to estimate the Lipschitz constant for neural networks \citep{scaman2018lipschitz}. In our empirical experiments, we see that when $x^r$ has the same classification as $x_\phi$, this constraint is automatically satisfied via a direct path.

% change this a bit - we are interested to find images in the pixel space that would map to particular points in the feature space.

\section{Convex hull of training set in feature space} \label{sec:fs_hull}

We now turn our attention to geometric properties of training and testing set in the feature space. Mainly, we investigate the geometry of testing samples with respect to the convex hull of training set. Using equation~\eqref{eq:map2f}, we can map all training samples to $\Phi$ and form their convex hull. $\mathcal{H}^{tr}_\phi$ denotes the convex hull of training set in $\Phi$ while \hull denotes the convex hull of training set in the pixel space. Furthermore, projection of $x$ to \hull is denoted by $x^h$, and projection of $x_\phi$ to $\mathcal{H}^{tr}_\phi$ is denoted by $x^h_\phi$.

It is reported that for standard image classification datasets, testing samples are entirely outside $\mathcal{H}^{tr}$ and $\mathcal{H}^{tr}_\phi$. As a result, a model has to extrapolate in order to classify testing samples \citep{yousefzadeh2021hull,balestriero2021learning}. Here, we study the extent of such extrapolation in the feature space and investigate its implications for the pixel space. Particularly, for a given $x$ and its corresponding $x^h_\phi$, we would like to find the least changes in $x$ that would directly map it to $x^h_\phi$. Moreover, using the formulations in previous sections, we will investigate the decision boundaries of the model in feature space with respect to the $\mathcal{H}^{tr}_\phi$, as presented in numerical experiments. Before that, we briefly review the computations necessary to project a point to a convex hull.

% How robust is the feature space that deep neural networks learn from images?

% Here, we are interested to gain an understanding of the convex hull of training set in the feature space. Particularly, for a given $x$ and its corresponding $x^h_\phi$, we would like to find the least changes in $x$ that would directly map it to $x^h_\phi$. This motivates us to formulate and solve an inverse problem. But, let's first review the process in which we compute the $x^h_\phi$ for a given $x$.

\subsection{Projecting a query point to a convex hull}

In the feature space, as in the pixel space, projecting a query point to a convex hull can be performed by solving a convex optimization problem. In previous work, we have provided an algorithm to solve it faster than off-the-shelf algorithms \citep{yousefzadeh2021sketching}.% Here, we briefly review the formulation. \citep{yousefzadeh2021sketching}

Given a point in the feature space, $x_\phi$, we would like to find the closest point to it on the $\mathcal{H}^{tr}_\phi$. Distance can be measured using any desired norm. Here, we use the 2-norm distance and minimize it via the objective function
\begin{equation} \label{eq:hull_obj}
	\min_{x^h_\phi} \| x^h_\phi - x_\phi \|_2^2
\end{equation}
Our first constraint relates the solution to the samples in training set
\begin{equation} \label{eq:hull_c1}
  x^h_\phi = \alpha \mathcal{D}_\phi,
\end{equation}
where $\mathcal{D}_\phi$ is the training set, in the feature space, formed as a matrix where rows represent $n$ samples and columns represent $d_\phi$ features. The other two constraints ensure that $x^h_\phi$ belongs to the convex hull of $\mathcal{D}_\phi$.
\begin{equation} \label{eq:hull_c2}
  \alpha  \mathbbm{1}_{n,1} = 1,
\end{equation}
% \vspace{-.5cm}
\begin{equation} \label{eq:hull_c3}
  0 \leq \alpha.
\end{equation}
Minimizing the objective function \eqref{eq:hull_obj} subject to constraints~\eqref{eq:hull_c1}-\eqref{eq:hull_c3} will lead to the point on $\mathcal{H}^{tr}_\phi$ closest to $x_\phi$. Since our optimization problem is convex, there is guarantee to find its solution. We denote this projection with
\begin{equation} \label{eq:project_hull}
  x_\phi^h = \mathcal{P}^h(x_\phi,\mathcal{H}^{tr}_\phi),
\end{equation}
while distance to $\mathcal{H}^{tr}_\phi$ is denoted by
\begin{equation} \label{eq:dist_hull}
    d_\phi^{h}(x_\phi) = \| x_\phi - x_\phi^h \|_2.
\end{equation}
Using the optimization problem formulated in Section~\ref{sec:map2point}, we may map $x^h_\phi$ back to the pixel space.

% Next, we explain how $x^h_\phi$ may be mapped back to the pixel space.

% Using the equation \eqref{eq:map2f}, we first map $x$ to the feature space, and then we solve the optimization problem defined by equations \eqref{}-\eqref{} to obtain $x$. 

\section{Numerical experiments} \label{sec:results}

We first investigate a single image of CIFAR-10 dataset \citep{krizhevsky2009learning} in detail and from different perspectives. Later in Section~\ref{sec:results_cif10_trends}, we report the larger trends in this dataset.

\subsection{Insights about one image} \label{sec:results_one}

\begin{figure}[h]
\floatconts
  {fig:fs_cif10_t1}
  {    \vspace{-.5cm}
\caption{First testing sample in CIFAR-10 dataset {\bf (a)} in pixel space. {\bf (b)} mapping of $x$ to the feature space in the last hidden layer of a trained model.}}
  {%
    \subfigure[]{\label{fig:fs_cif10_t1_orig}%
      \includegraphics[width=0.12\linewidth]{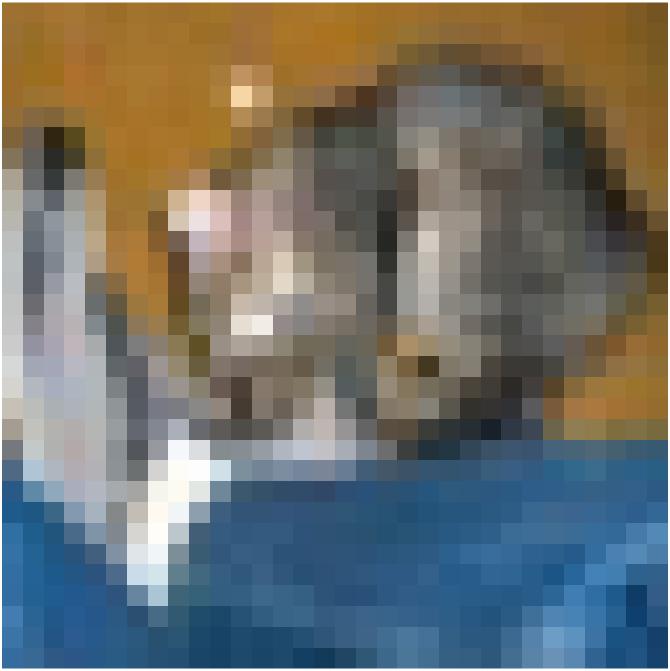}} \quad
    \subfigure[]{\label{fig:fs_cif10_t1_fs}%
      \includegraphics[width=0.13\linewidth]{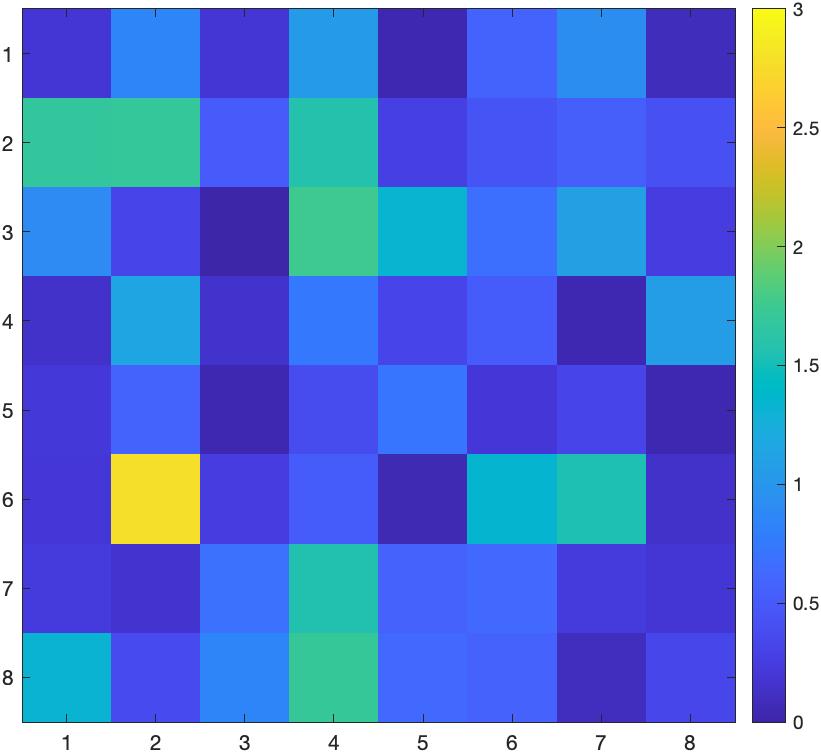}}}
\end{figure}

Let us consider $x$ to be the first testing sample of dataset shown in Figure~\ref{fig:fs_cif10_t1_orig}. Our model is a standard pre-trained model described in Section~\ref{sec:boundaries_fs} and available at the link in footnote~\ref{fn:model}. We map this image to the feature space of the model to obtain $x_\phi$. Since $x_\phi$ has 64 elements, we can plot it as an 8 by 8 image:% as shown in Figure~\ref{fig:fs_cif10_t1_fs}.

% \begin{figure}[h]
%   \centering
%   \includegraphics[width=0.2\linewidth]{figures/cif10_te1orig.jpg}
%   \caption{First testing sample in CIFAR-10 dataset, input $x$. and $x_\phi$}
%   \label{fig:fs_cif10_t1_orig}
% \end{figure}

% \begin{figure}[h]
%      \centering
%      \begin{subfigure}[b]{0.15\textwidth}
%          \centering
%          \includegraphics[width=\textwidth]{figures/cif10_te1orig.jpg}
%          \caption{$x$}
%          \label{fig:fs_cif10_t1_orig}
%      \end{subfigure}
%      \quad
%      \begin{subfigure}[b]{0.164\textwidth}
%          \centering
%          \includegraphics[width=\textwidth]{figures/cif10_te1_fs.jpg}
%          \caption{$x_\phi$}
%          \label{fig:fs_cif10_t1_fs}
%      \end{subfigure}
%         \caption{First testing sample in CIFAR-10 dataset {\bf (a)} in pixel space. {\bf (b)} mapping of $x$ to the feature space in the last hidden layer of a trained model.}
%         \label{fig:fs_cif10_t1}
% \end{figure}

\vspace{-.5cm}

{\bf Decision boundaries in the feature space.}
We first investigate the decision boundaries of the model in the vicinity of $x_\phi$. Classification of the model for this image is Cat. Table~\ref{tbl:ex1_dist_boundaries} shows the margin of $x_\phi$ to each of the other 9 classes.

\begin{table}[h]
  \caption{Distance to decision boundaries of each class in the feature space $\Phi$ (sample in Figure~\ref{fig:fs_cif10_t1})}
  \label{tbl:ex1_dist_boundaries}
  \centering
  \begin{tabular}{ccccccccccc}
    \toprule
    % \multicolumn{2}{c}{Part}                   \\
    % \cmidrule(r){1-2}
    Class   & airplane     & car & bird & cat & deer & dog & frog & horse & ship & truck\\
    $j$ &  1     & 2    & 3     & 4  & 5 &  6 & 7 & 8 & 9 & 10  \\
    \midrule
    $d^{f(4,j)}_\phi$ & 3.498  & 3.266 & 2.546  & - & 3.087 & 2.629 & 2.711 & 3.805 & 3.494 & 3.849 \\
    \bottomrule
\end{tabular}
\end{table}

% Based on the closest flip point, which is with class bird, we have $d = $ for this image, defining the ball $\mathcal{B}$ centered at $x_\phi$ with radius .

\begin{figure}[h]
\floatconts
  {fig:fs_cif10_t1_flip}
  {    \vspace{-.5cm}
\caption{{\bf (a)} Closest flip point in $\Phi$ for image in Figure~\ref{fig:fs_cif10_t1}, {\bf (b)} difference between the closest flip point and $x_\phi$, {\bf (c)}~image that would directly map to $x_\phi^{f(4,3),c}$}}
  {%
    \subfigure[${\scriptstyle x_\phi^{f(4,3),c}}$]{\label{fig:fs_cif10_t1_fs_flip}%
      \includegraphics[width=0.14\linewidth]{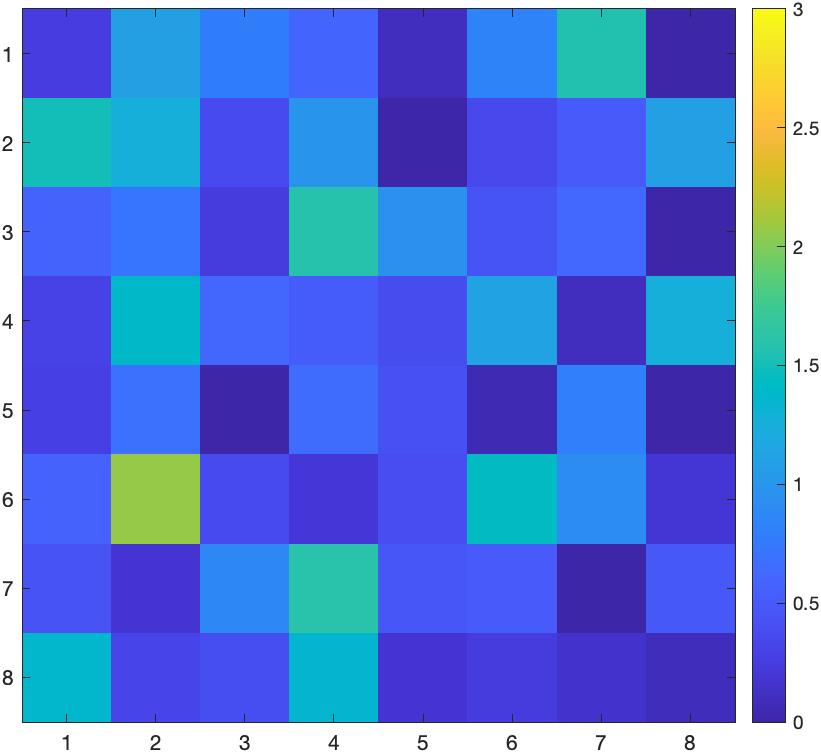} \vspace{.5cm}} \quad
    \subfigure[${\scriptstyle | x_\phi^{f(4,3),c} - x_\phi |}$]{\label{fig:fs_cif10_t1_fs_flip_diff}%
      \includegraphics[width=0.14\linewidth]{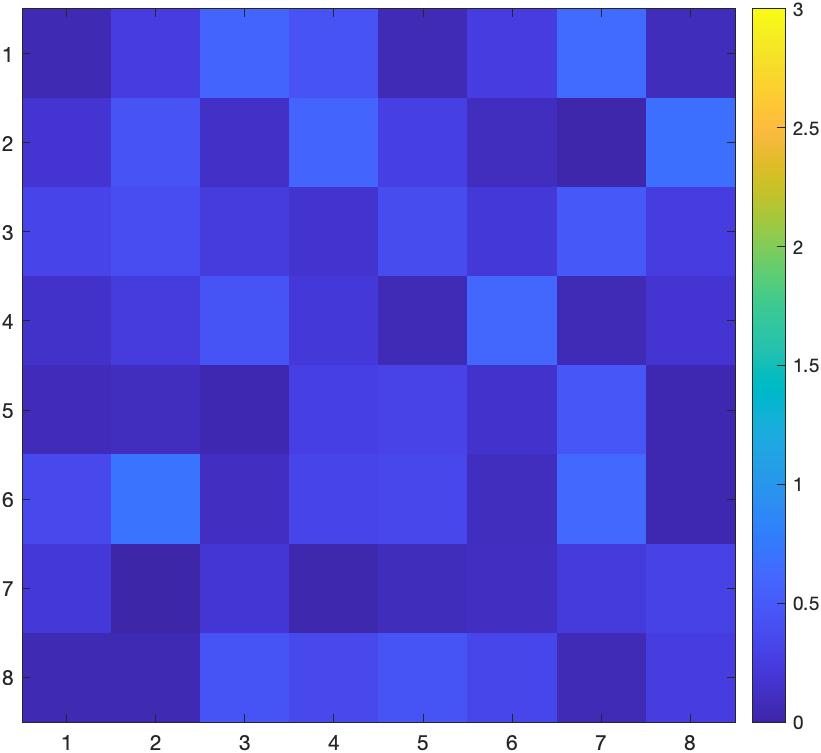}} \quad
    \subfigure[$x^{\Omega \rightarrow x_\phi^{f(4,3),c}}$]{\label{fig:fs_cif10_t1_fs_flip_p}%
      \includegraphics[width=0.12\linewidth]{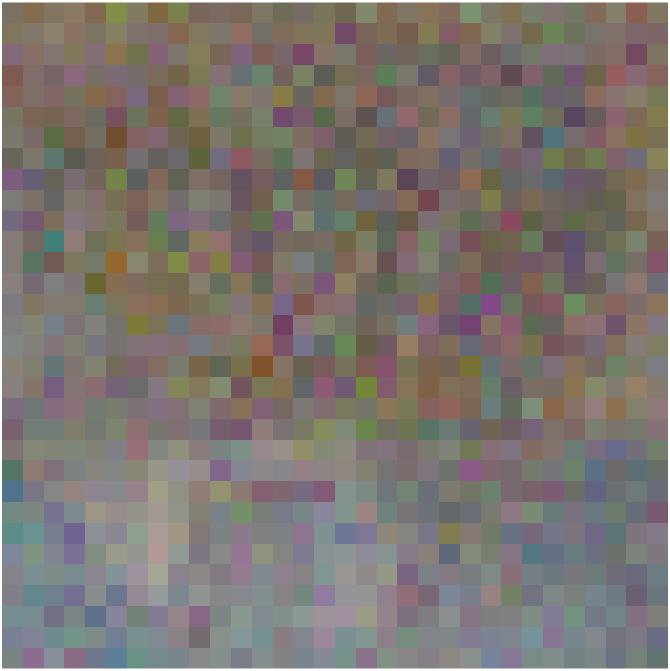}}}
\end{figure}

{\bf Closest flip point and $\mathcal{B}(x_\phi)$.} The flip point closest to $x_\phi$ is with the class bird, distanced 2.546 from it (measured in L2 norm in the 64-dimensional feature space). This flip point is depicted in Figure~\ref{fig:fs_cif10_t1_fs_flip}, and its distance to $x_\phi$ defines the radius of $\mathcal{B}(x_\phi)$. Any point in feature space that is a member of $\mathcal{B}(x_\phi) \cap \Omega_\phi$ (i.e., closer than 2.546 to $x_\phi$) is guaranteed to be classified as Cat by the model. Moreover, Lipschitz constant for the feature space is 6.122, the largest singular value of $W_\phi$, enabling us to study this space with clarity. Intriguingly, we see that 437 training samples and 69 testing samples are actually inside the $\mathcal{B}(x_\phi)\cap \Omega_\phi$ centered at $x_\phi$. % Figure~\ref{fig:} shows 4 of these samples. 
We then solve the optimization problem defined by equations~\eqref{eq:point_c1}-\eqref{eq:point_obj} to find the image in the pixel space that would map to this specific flip point, obtaining the image shown in Figure~\ref{fig:fs_cif10_t1_fs_flip_p}. This image can be considered the closest adversarial example in $\Phi$, however, in the pixel space, it looks very different from the original image.

% \begin{figure}[h]
%      \centering
%       \begin{subfigure}[b]{0.164\textwidth}
%          \centering
%          \includegraphics[width=\textwidth]{figures/cif10_te1_fs_cflip.jpg}
%          \caption{${\scriptstyle x_\phi^{f(4,3),c}}$}
%          \label{fig:fs_cif10_t1_fs_flip}
%      \end{subfigure}
%      \quad
%     \begin{subfigure}[b]{0.164\textwidth}
%          \centering
%          \includegraphics[width=\textwidth]{figures/cif10_te1_fs_cflip_diff.jpg}
%          \caption{${\scriptstyle | x_\phi^{f(4,3),c} - x_\phi |}$}
%          \label{fig:fs_cif10_t1_fs_flip_diff}
%      \end{subfigure}
%      \quad
%      \begin{subfigure}[b]{0.15\textwidth}
%          \centering
%          \includegraphics[width=.97\textwidth]{figures/cif10_te1_fs_flip_p.jpg}\vspace{-.1cm}
%          \caption{$x^{\Omega \rightarrow x_\phi^{f(4,3),c}}$}
%          \label{fig:fs_cif10_t1_fs_flip_p}
%      \end{subfigure}
%         \caption{{\bf (a)} Closest flip point in $\Phi$ for image in Figure~\ref{fig:fs_cif10_t1}, {\bf (b)} difference between the closest flip point and $x_\phi$, {\bf (c)}~image that would directly map to $x_\phi^{f(4,3),c}$}
%         \label{fig:fs_cif10_t1_flip}
% \end{figure}

\vspace{-.5cm}

\begin{figure}[h]
\floatconts
  {fig:fs_cif10_t1_hull_p}
  {    \vspace{-.7cm}
\caption{First testing sample in CIFAR-10 dataset {\bf (a)} in pixel space, {\bf (b)} its projection to \hull, {\bf (c)} their difference.}\vspace{-.5cm}}
  {%
    \subfigure[$x$]{\label{fig:fs_cif10_t1_orig_repeat}%
      \includegraphics[width=0.12\linewidth]{figures/cif10_te1orig.jpg}} \quad
    \subfigure[$x^h$]{\label{fig:fs_cif10_t1_hull}%
      \includegraphics[width=0.12\linewidth]{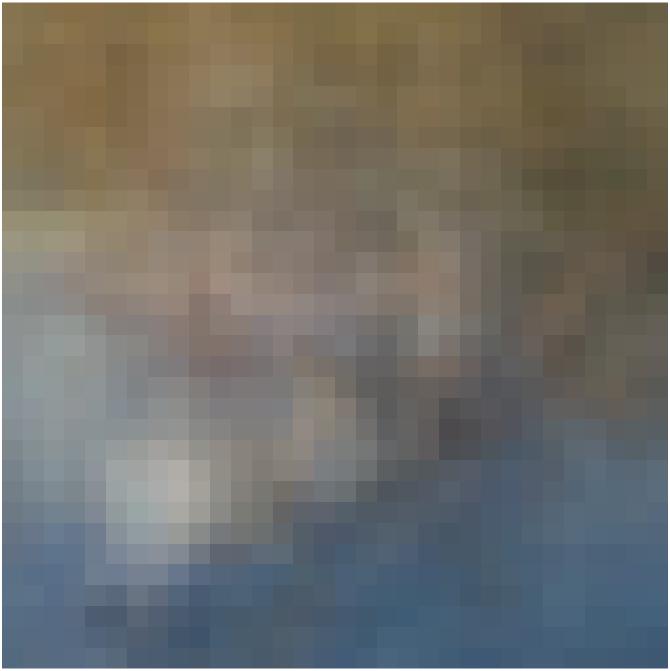}} \quad
    \subfigure[$| x^h - x |$]{\label{fig:fs_cif10_t1_hull_diff}%
      \includegraphics[width=0.12\linewidth]{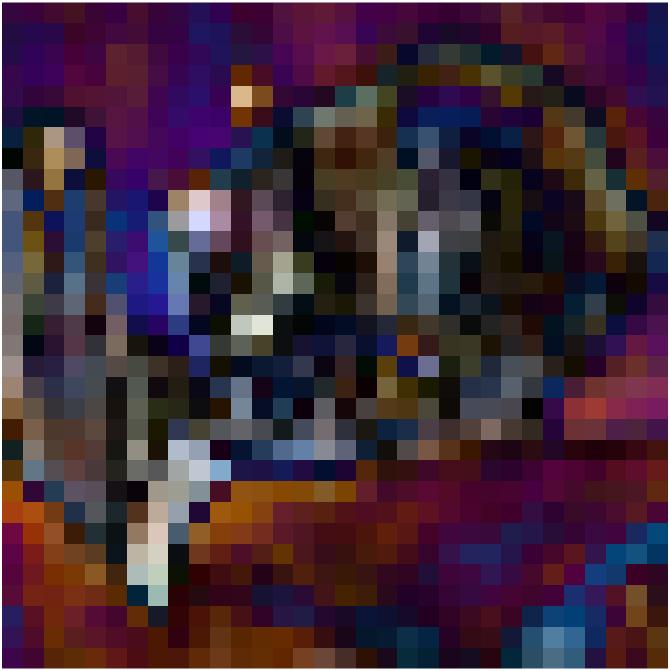}}}
\end{figure}

{\bf Convex hull of training set in feature space.} The fact that some training samples are members of $\mathcal{B}(x_\phi)\cap \Omega_\phi$ implies that the convex hull of the training set overlaps with $\mathcal{B}(x_\phi)\cap \Omega_\phi$. Let us remember that this testing sample, as well as all other testing samples of this dataset, are outside the convex hull of training set, both in pixel space and in feature space. However, geometric arrangements are different in the feature space. In the pixel space, usually, decision boundaries are very close to both training and testing samples. It is known that adversarial examples, i.e., close-by images on the other side of decision boundaries, are so similar to original images that their differences are not easily detectable by human eye. At the same time, in the pixel space, convex hull of training set is rather far from images, and images have to visibly change to reach their \hull. See, for example, Figure~\ref{fig:fs_cif10_t1_hull} for the projection of our first testing sample to the convex hull of the training set in the pixel space, and notice that the image has considerably changed while changes are related to the object of interest as shown in Figure~\ref{fig:fs_cif10_t1_hull_diff}.% Model classifies the projection as .

% \begin{figure}[h]
%      \centering
%      \begin{subfigure}[b]{0.15\textwidth}
%          \centering
%          \includegraphics[width=\textwidth]{figures/cif10_te1orig.jpg}
%          \caption{$x$}
%          \label{fig:fs_cif10_t1_orig_repeat}
%      \end{subfigure}
%      \quad
%      \begin{subfigure}[b]{0.15\textwidth}
%          \centering
%          \includegraphics[width=\textwidth]{figures/cif10_te1_hull.jpg}
%          \caption{$x^h$}
%          \label{fig:fs_cif10_t1_hull}
%      \end{subfigure}
%      \quad
%      \begin{subfigure}[b]{0.15\textwidth}
%          \centering
%          \includegraphics[width=\textwidth]{figures/cif10_te1_hull_diff.jpg}
%          \caption{$| x^h - x |$}
%          \label{fig:fs_cif10_t1_hull_diff}
%      \end{subfigure}
%         \caption{First testing sample in CIFAR-10 dataset {\bf (a)} in pixel space, {\bf (b)} its projection to \hull, {\bf (c)} their difference.}
%         \label{fig:fs_cif10_t1_hull_p}
% \end{figure}

\begin{figure}[h]
\floatconts
  {fig:fs_cif10_t1_hull_fs}
  {    \vspace{-.7cm}
\caption{{\bf (a)} Projection of $x_\phi$ to convex hull of training set in feature space, {\bf (b)} difference with $x_\phi$, {\bf (c)} image that would directly map to $x_\phi^h$.} \vspace{-.7cm}}
  {%
    \subfigure[$x_\phi^h$]{\label{fig:fs_cif10_t1_fs_hull}%
      \includegraphics[width=0.14\linewidth]{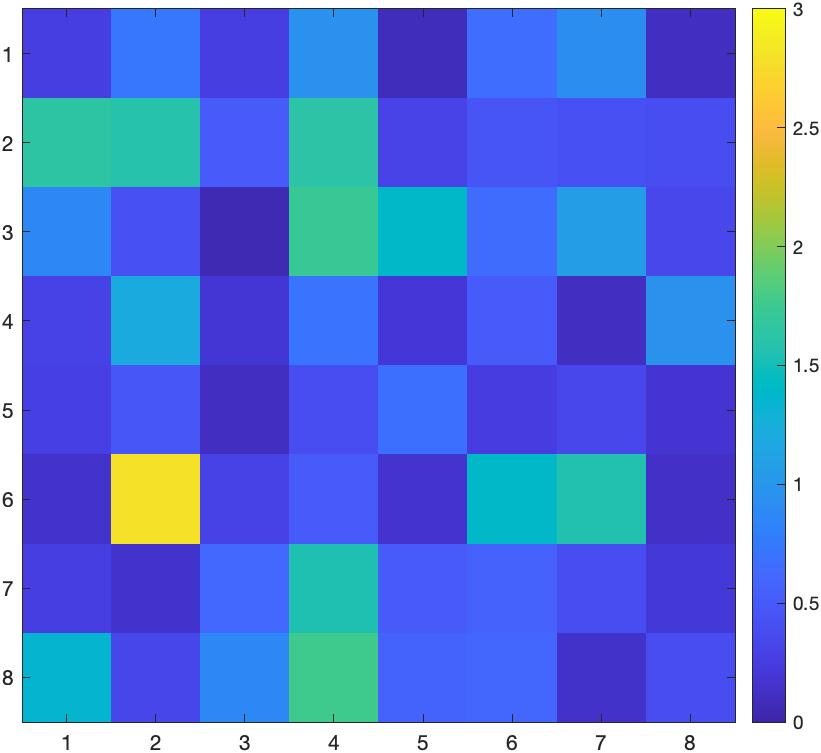}} \quad
    \subfigure[$| x_\phi^h - x_\phi |$]{\label{fig:fs_cif10_t1_fs_hull_diff}%
      \includegraphics[width=0.14\linewidth]{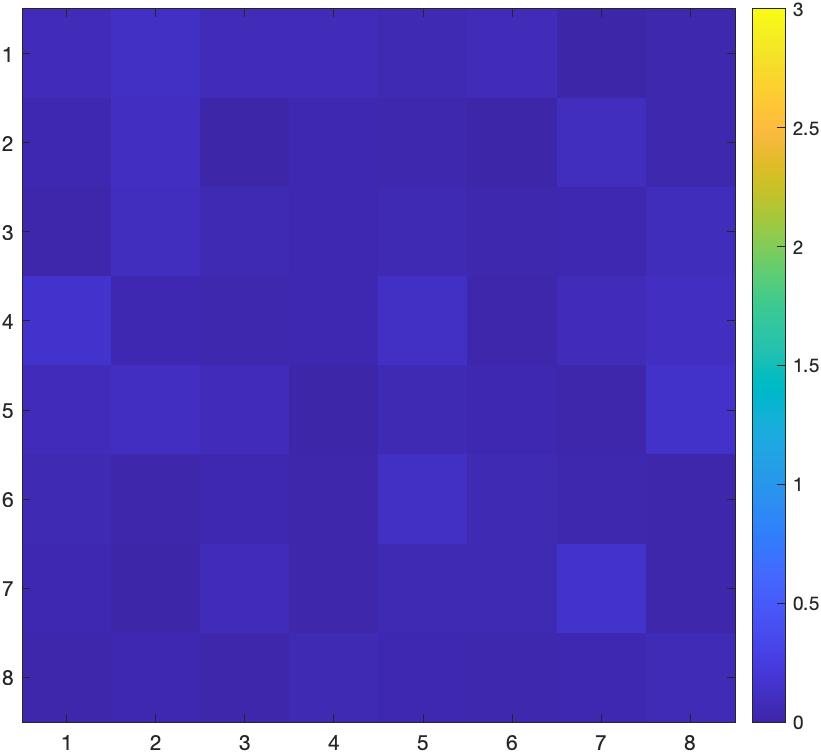}} \quad
    \subfigure[$x^{\Omega \rightarrow x_\phi^h}$]{\label{fig:fs_cif10_t1_fs_hull_p}%
      \includegraphics[width=0.12\linewidth]{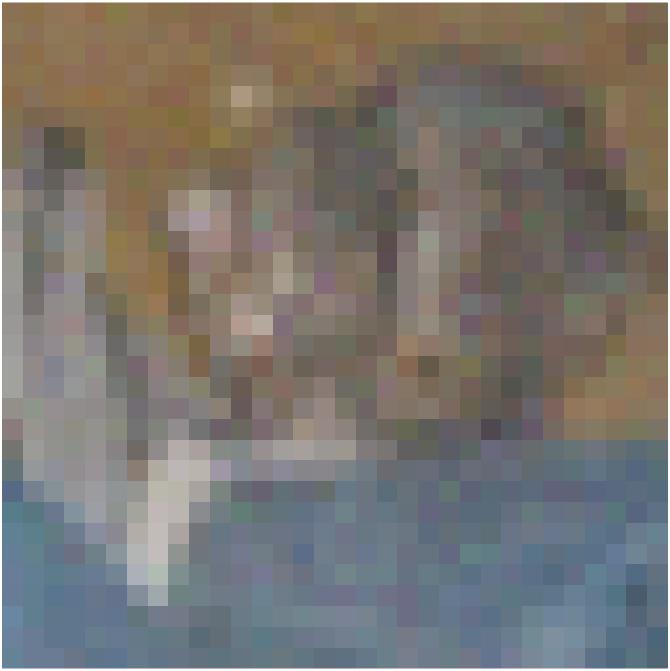}}}
\end{figure}

In feature space, however, this order is reversed, i.e., convex hull of training set is closer to the sample compared to decision boundaries. Figure~\ref{fig:fs_cif10_t1_fs_hull} shows the projection of our testing sample to $\mathcal{H}_\phi^{tr}$ using equation~\eqref{eq:project_hull}. This point is distanced 0.508 from $x_\phi$, smaller than the 2.546 distance to closest decision boundary in feature space. Notice that the corresponding image in Figure~\ref{fig:fs_cif10_t1_fs_hull_p}, derived from equations in Section~\ref{sec:map2point}, looks more similar to the original image compared to the closest image on the decision boundary shown in Figure~\ref{fig:fs_cif10_t1_fs_flip_p} and also the projection in the pixel space shown in Figure~\ref{fig:fs_cif10_t1_hull}. Hence, in the feature space, testing sample is more closely related to the convex hull of training set.

% \begin{figure}[h]
%      \centering
%      \begin{subfigure}[b]{0.164\textwidth}
%          \centering
%          \includegraphics[width=\textwidth]{figures/cif10_te1_fs_hull.jpg}
%          \caption{$x_\phi^h$}
%          \label{fig:fs_cif10_t1_fs_hull}
%      \end{subfigure}
%      \quad
%      \begin{subfigure}[b]{0.164\textwidth}
%          \centering
%          \includegraphics[width=\textwidth]{figures/cif10_te1_fs_hull_diff.jpg}
%          \caption{$| x_\phi^h - x_\phi |$}
%          \label{fig:fs_cif10_t1_fs_hull_diff}
%      \end{subfigure}
%      \quad
%      \begin{subfigure}[b]{0.15\textwidth}
%          \centering
%          \includegraphics[width=\textwidth]{figures/cif10_te1_fs_hull_p.jpg}
%          \caption{$x^{\Omega \rightarrow x_\phi^h}$}
%          \label{fig:fs_cif10_t1_fs_hull_p}
%      \end{subfigure}
%         \caption{{\bf (a)} projection of $x_\phi$ to convex hull of training set in feature space, {\bf (b)} difference with $x_\phi$, {\bf (c)} image that would directly map to $x_\phi^h$.}
%         \label{fig:fs_cif10_t1_hull_fs}
% \end{figure}

% \vspace{-.5cm}

% Put differently, our standard pre-trained model has drawn a decision boundary separating this testing sample from the training set, and as a result, when we project the image to the training set, its classification changes. On the other hand, the relationship between the testing sample and the training set is more intimate in $\Phi$, as there is no decision boundary separating this sample from the relevant portion of training set. This image is no anomaly and these trends persist for most testing samples as we shall see.

% Next, we investigate this sample in relation to the convex hull of training set. This sample is outside the $\mathcal{H}^{tr}$ and $\mathcal{H}^{tr}_\phi$ as is the case for all testing samples of this dataset.

\begin{figure}[h]
\floatconts
  {fig:cif10_te1_supp}
  {    \vspace{-.7cm}
\caption{Images that form the point on the convex hull of training set, closest to image~\ref{fig:fs_cif10_t1_orig}.}}
  {%
    \subfigure[Support images in pixel space]{\label{fig:cif10_te1_supp_pixel}%
      \includegraphics[width=0.1\linewidth]{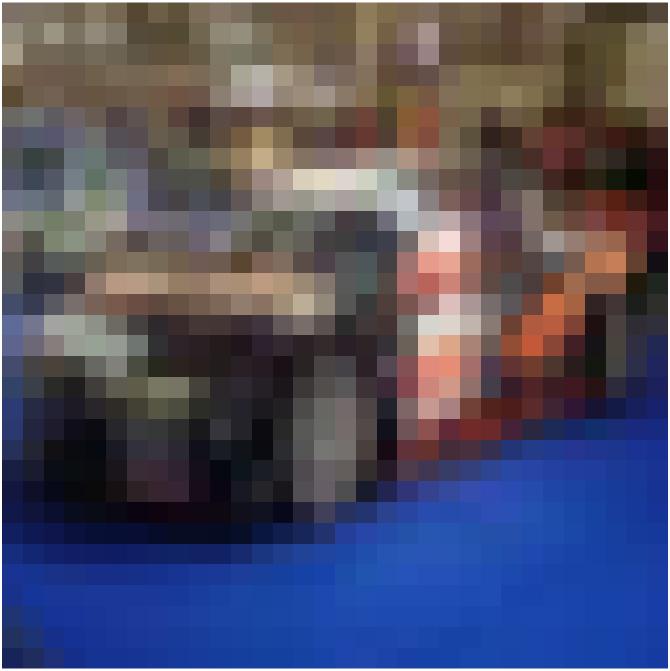} 
      \includegraphics[width=0.1\linewidth]{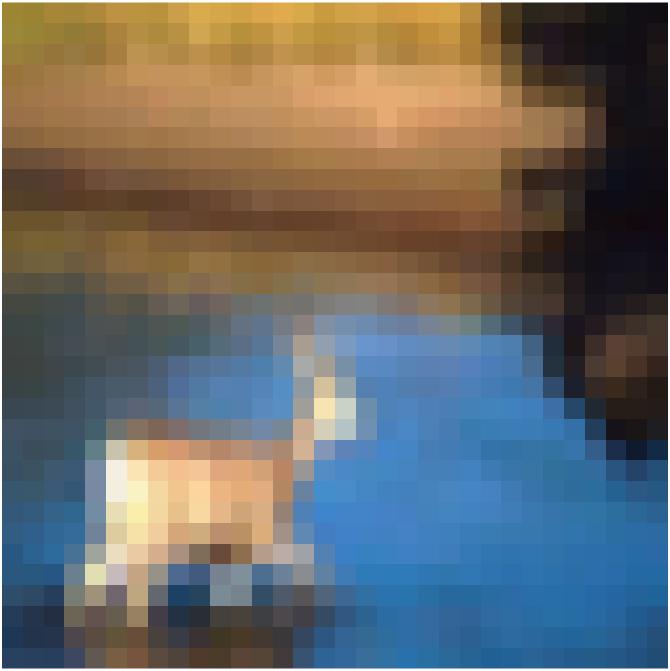} 
      \includegraphics[width=0.1\linewidth]{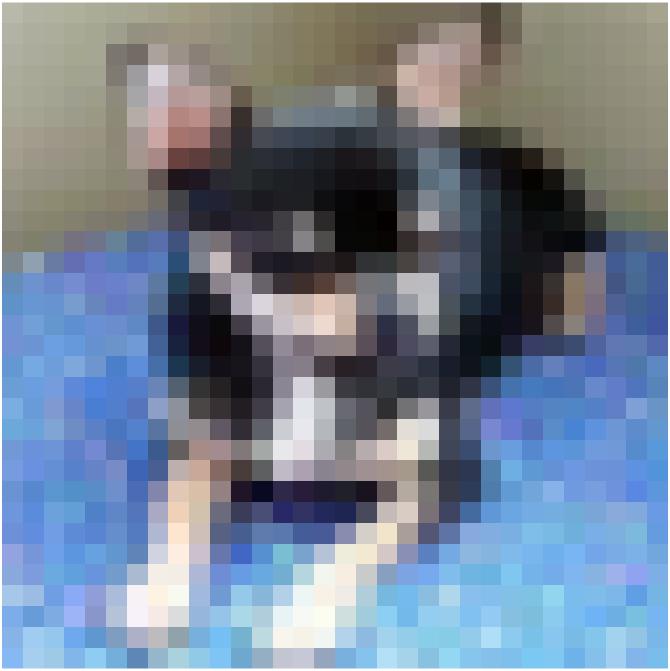} 
      \includegraphics[width=0.1\linewidth]{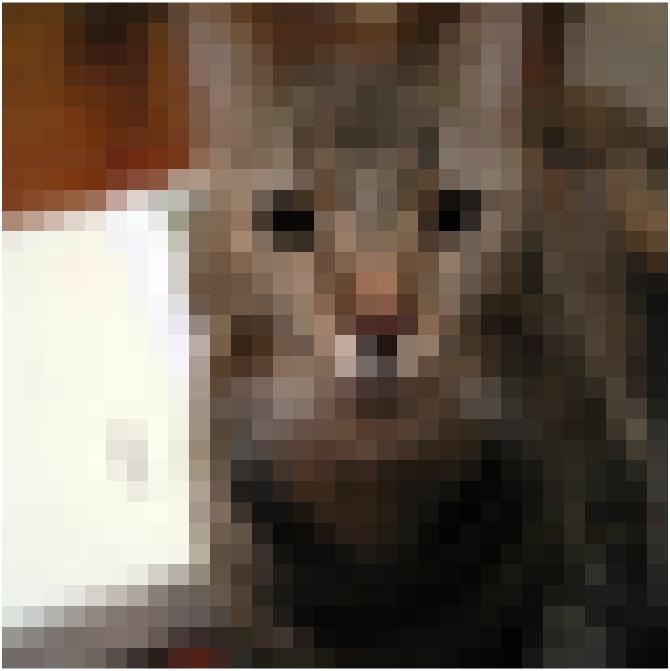}}\qquad \quad 
    \subfigure[Support images in feature space]{\label{fig:cif10_te1_supp_fs}%
      \includegraphics[width=0.1\linewidth]{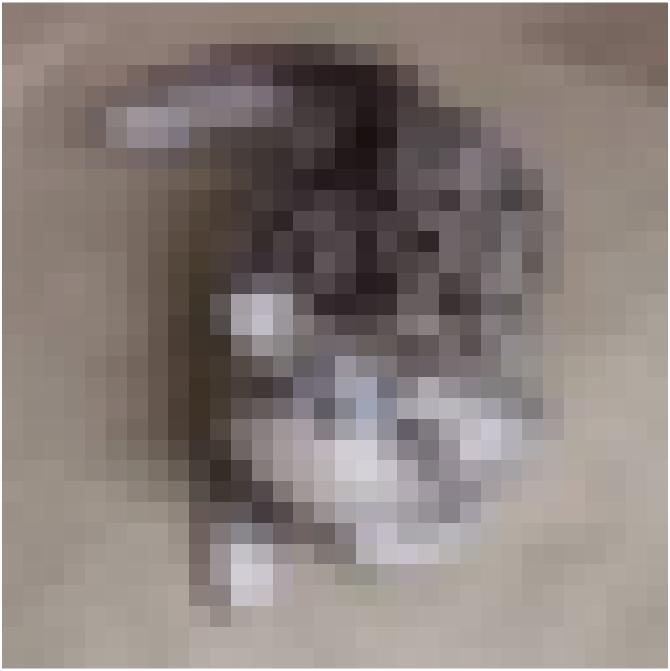}
      \includegraphics[width=0.1\linewidth]{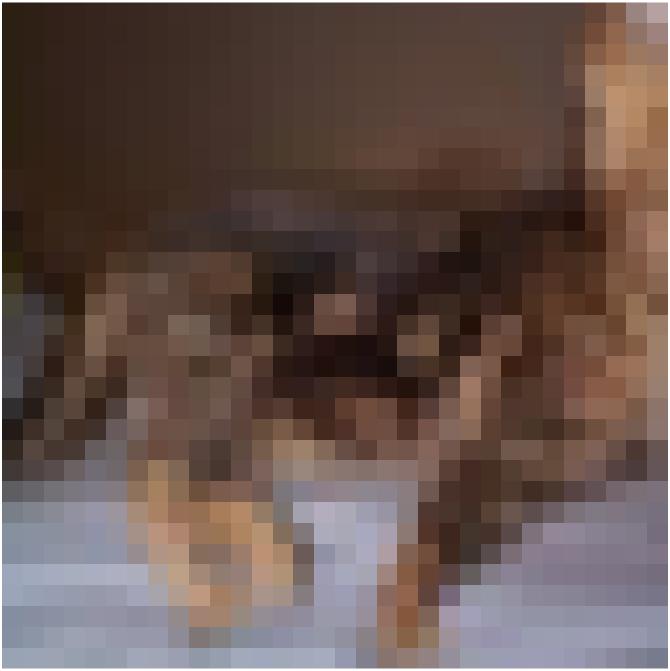}
      \includegraphics[width=0.1\linewidth]{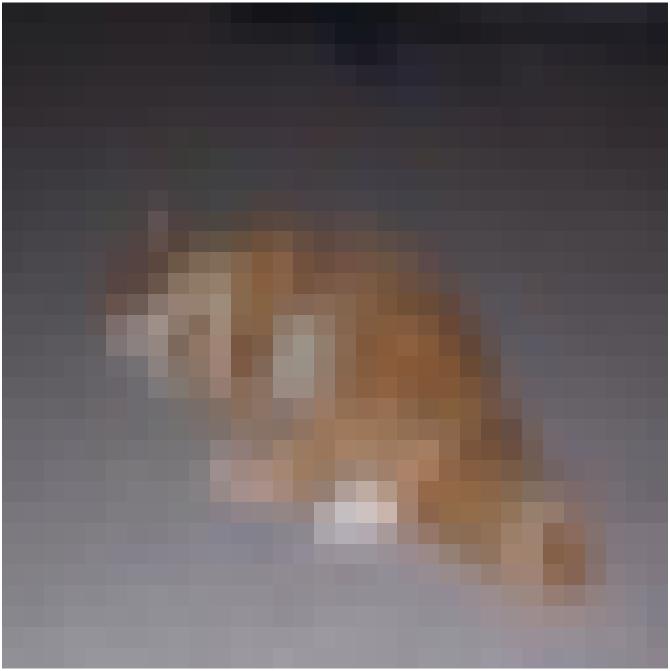}
      \includegraphics[width=0.1\linewidth]{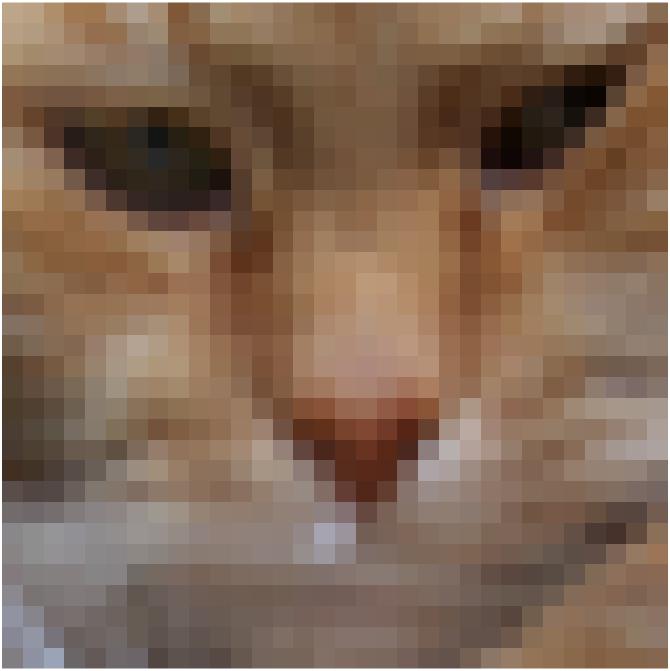}}}
\end{figure}

{\bf Support in the training set.} Let us now look at training images that participate in the convex combination leading to $x^h$ and $x_\phi^h$. Figure~\ref{fig:cif10_te1_supp_pixel} shows four images with largest $\alpha$ coefficients that contribute to the convex hull projection in pixel space, shown in Figure~\ref{fig:fs_cif10_t1_hull}. Coefficients refer to the optimization parameter $\alpha$ in equation~\eqref{eq:hull_c1}. Note that only one of these images is from the Cat class while others are from the classes of Automobile, Deer, and Dog.

% \begin{figure}[h]
%      \centering
%      \begin{subfigure}[b]{0.65\textwidth}
%          \centering
%          \includegraphics[width=.2\textwidth]{figures/cif10_te1supp_p_tr44530.jpg} \:
%          \includegraphics[width=.2\textwidth]{figures/cif10_te1supp_p_tr12060.jpg} \:
%          \includegraphics[width=.2\textwidth]{figures/cif10_te1supp_p_tr3006.jpg} \:
%          \includegraphics[width=.2\textwidth]{figures/cif10_te1supp_p_tr5089.jpg}
%          \caption{Support images in pixel space}
%          \label{fig:cif10_te1_supp_pixel}
%      \end{subfigure}
%      \quad
%      \begin{subfigure}[b]{0.65\textwidth}
%          \centering
%          \includegraphics[width=.2\textwidth]{figures/cif10_te1supp_fs_tr10183.jpg} \:
%          \includegraphics[width=.2\textwidth]{figures/cif10_te1supp_fs_tr31384.jpg} \:
%          \includegraphics[width=.2\textwidth]{figures/cif10_te1supp_fs_tr4109.jpg} \:
%          \includegraphics[width=.2\textwidth]{figures/cif10_te1supp_fs_tr7122.jpg}
%          \caption{Support images in feature space}
%          \label{fig:cif10_te1_supp_fs}
%      \end{subfigure}
%         \caption{Images that form the point on the convex hull of training set, closest to the first testing sample of CIFAR-10.}
%         \label{fig:cif10_te1_supp}
% \end{figure}

% \vspace{-.5cm}

Similarly, Figure~\ref{fig:cif10_te1_supp_fs} shows the training images with largest $\alpha$ coefficients supporting the projection of our image to the convex hull in feature space. These image are all from the Cat class, and the resulting image in the pixel space (Figure~\ref{fig:fs_cif10_t1_fs_hull_p}) looks more similar to the original image.

{\bf Images on the perimeter of $\mathcal{B}(x_\phi)$.}
We seek images in the pixel space that would map to the perimeter of $\mathcal{B}(x_\phi)\cap \Omega_\phi$ in the feature space. This is done by solving the optimization problem defined by equations~\eqref{eq:ball_c1}-\eqref{eq:ball_obj} using $r=2.546$ and with different reference points. To ensure images are on the perimeter, we change the inequality constraint of equation~\eqref{eq:ball_c1} to equality constraint. Finding images on the perimeter of $\mathcal{B}(x_\phi)\cap \Omega_\phi$ can be informative because it shows the extremes of $\mathcal{B}(x_\phi)\cap \Omega_\phi$. Resulting images are shown in Figure~\ref{fig:fs_cif10_te1_per} next to their reference points.

\begin{figure}[h]
\floatconts
  {fig:fs_cif10_te1_per}
  {    \vspace{-.7cm}
\caption{A variety of images in pixel space may map to the perimeter of $\mathcal{B}(x_\phi)$ for a particular image. The second image in each box is on the perimeter of $\mathcal{B}(x_\phi)$ for the first testing sample of CIFAR-10.}\vspace{-.5cm}}
  {%
  \fbox{
    \subfigure[$x_r$\vspace{.05cm}]{\label{fig:fs_cif10_tr19821}%
      \includegraphics[width=0.11\linewidth]{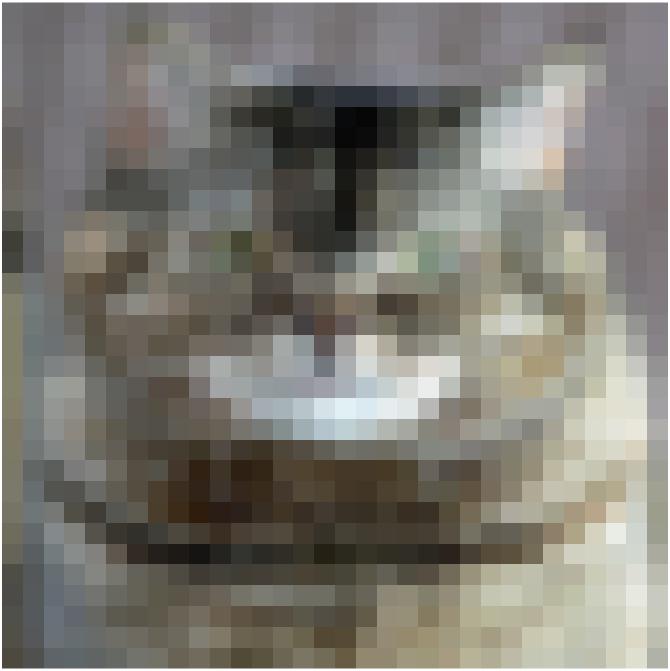}} \quad
    \subfigure[$x^{\Omega \rightarrow \mathcal{B}(x_\phi)}$]{\label{fig:fs_cif10_tr19821_per_te1}%
      \includegraphics[width=0.11\linewidth]{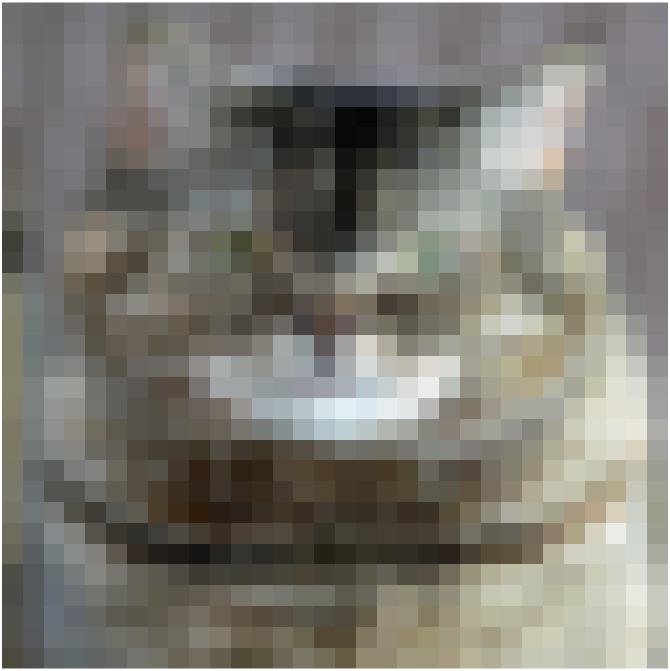}} \quad
    \subfigure[$\mathcal{N}_\phi(.)$ \vspace{.05cm}]{\label{fig:fs_cif10_tr19821_per_te1_fs}%
      \includegraphics[width=0.12\linewidth]{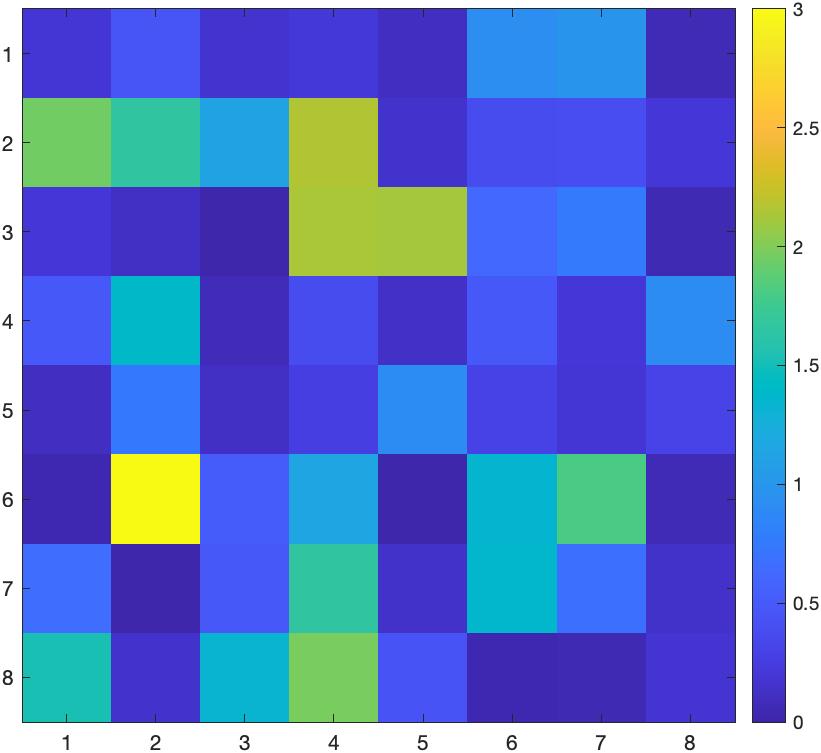}}}\qquad \qquad
\fbox{    \subfigure[$x_r$\vspace{.05cm}]{\label{fig:fs_cif10_tr2}%
      \includegraphics[width=0.11\linewidth]{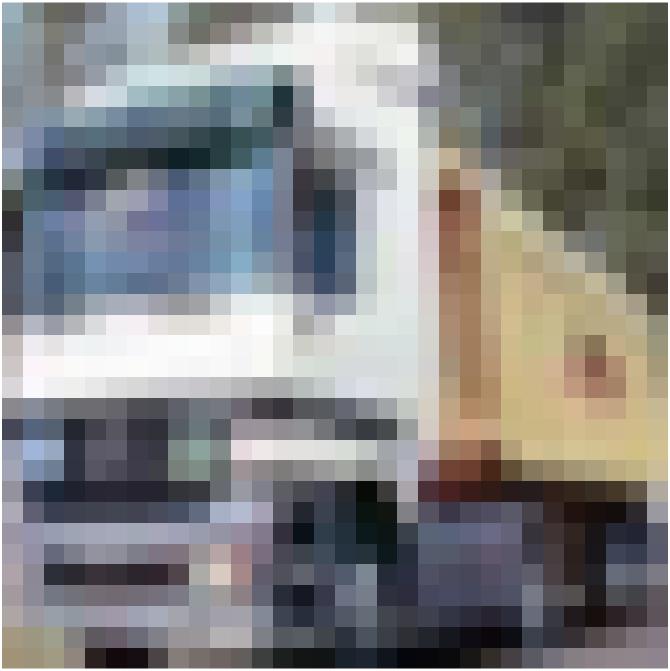}} \quad
    \subfigure[$x^{\Omega \rightarrow \mathcal{B}(x_\phi)}$]{\label{fig:fs_cif10_tr2_per_te1}%
      \includegraphics[width=0.11\linewidth]{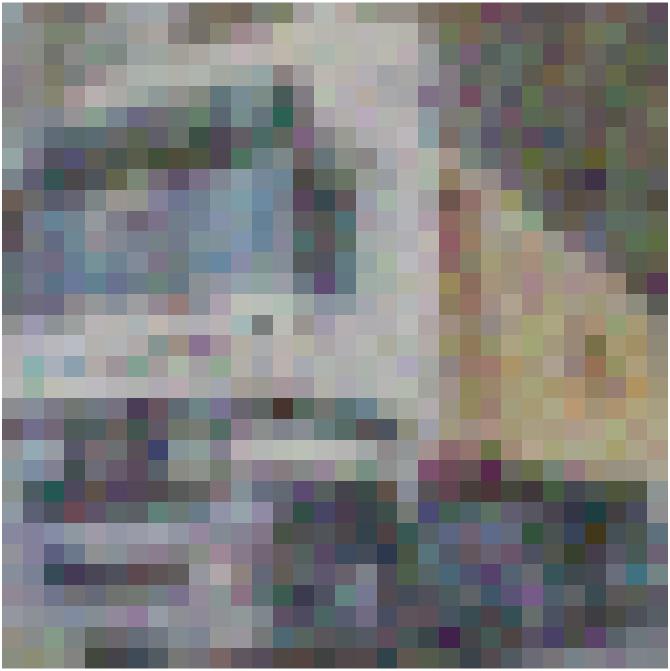}} \quad
    \subfigure[$\mathcal{N}_\phi(.)$ \vspace{.05cm}]{\label{fig:fs_cif10_tr2_per_te1_fs}%
      \includegraphics[width=0.12\linewidth]{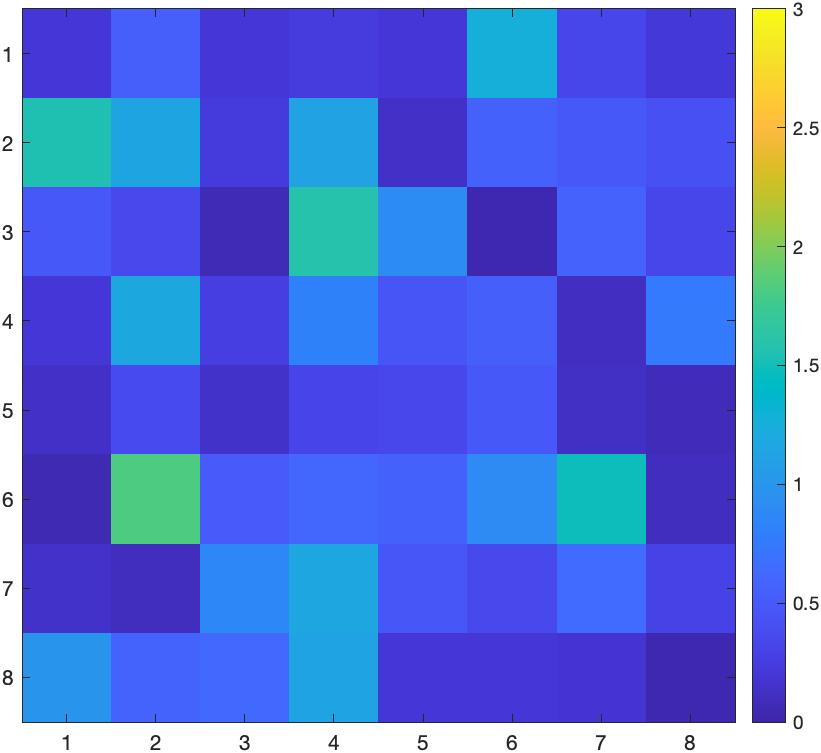}}}\\      
\fbox{    \subfigure[$x_r$\vspace{.05cm}]{\label{fig:fs_cif10_te2}%
      \includegraphics[width=0.11\linewidth]{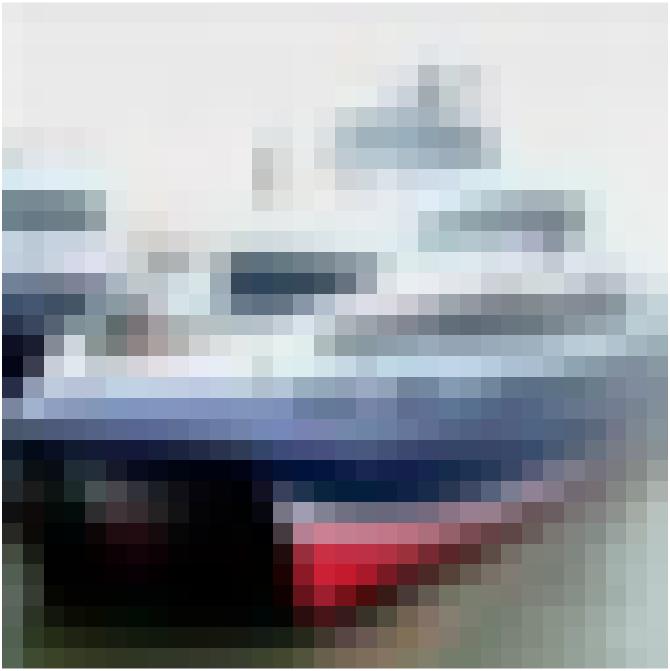}} \quad
    \subfigure[$x^{\Omega \rightarrow \mathcal{B}(x_\phi)}$]{\label{fig:fs_cif10_te2_per_te1}%
      \includegraphics[width=0.11\linewidth]{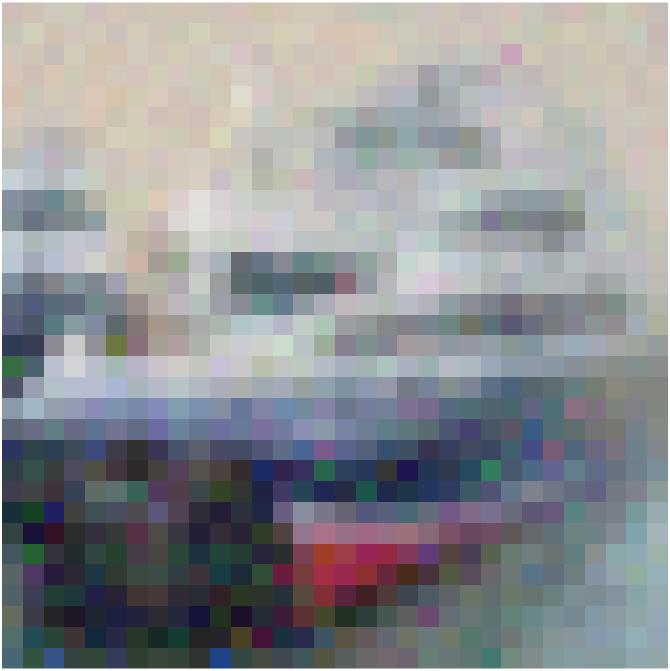}} \quad
    \subfigure[$\mathcal{N}_\phi(.)$ \vspace{.05cm}]{\label{fig:fs_cif10_te2_per_te1_fs}%
      \includegraphics[width=0.12\linewidth]{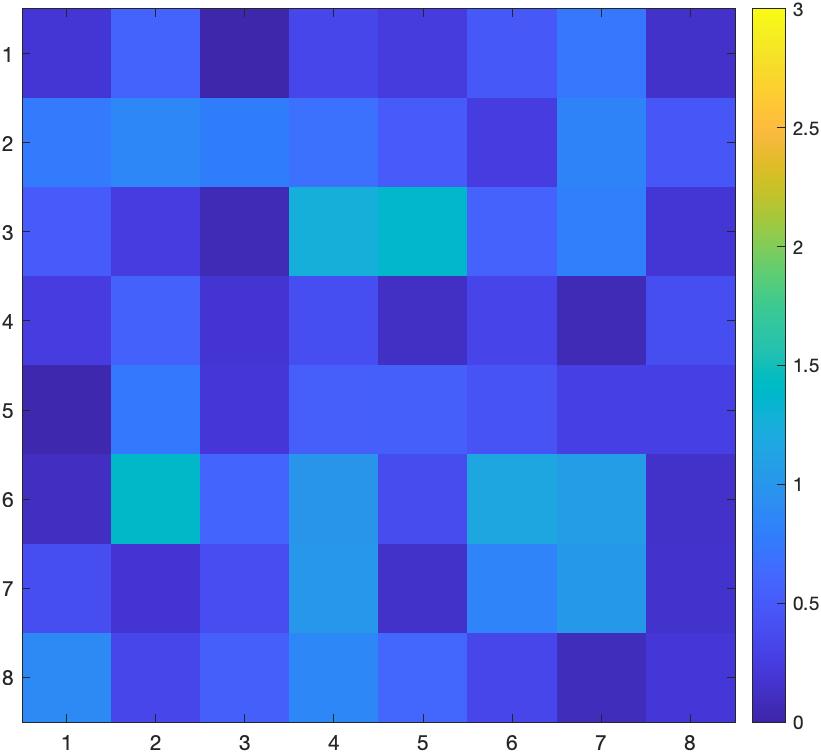}}}}
\end{figure}

\begin{figure}[h]
\floatconts
  {fig:cif10_te1_morph19821}
  {    \vspace{-.7cm}
\caption{Morphing between two images in the feature space. Distance is measured from the projection of $l$ in feature space. This entire path is inside the $\mathcal{B}(x_\phi)$ and classified as Cat.}\vspace{-.6cm}}
  {%
    \subfigure[2.54]{\label{fig:cif10_tr19821}%
      \includegraphics[width=0.10\linewidth]{figures/cif10_tr19821.jpg}} \:
    \subfigure[2.50]{\label{fig:fs_cif10_te1_path_tr19821_250}%
      \includegraphics[width=0.10\linewidth]{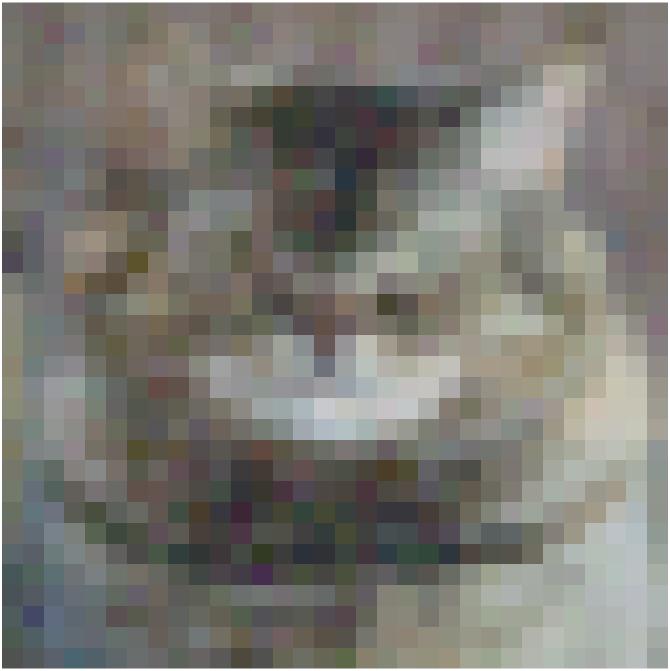}} \:
    \subfigure[2.40]{\label{fig:fs_cif10_te1_path_tr19821_240}%
      \includegraphics[width=0.10\linewidth]{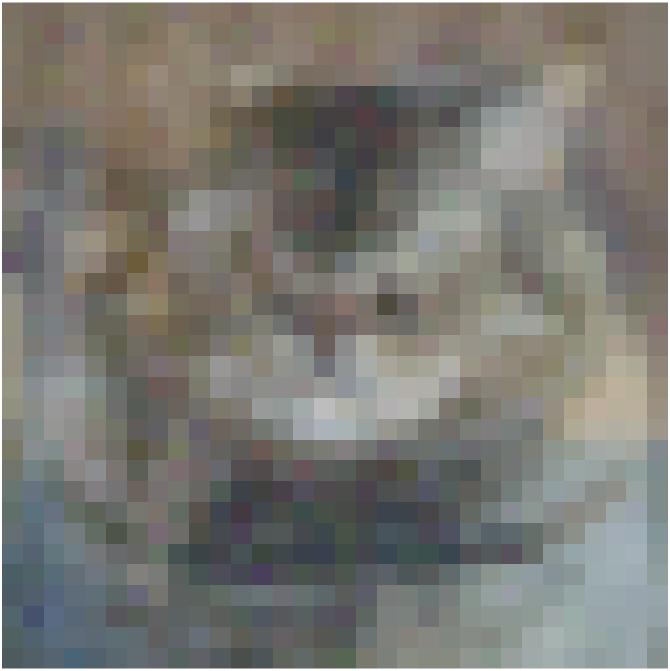}} \:
    \subfigure[2.25]{\label{fig:fs_cif10_te1_path_tr19821_225}%
      \includegraphics[width=0.10\linewidth]{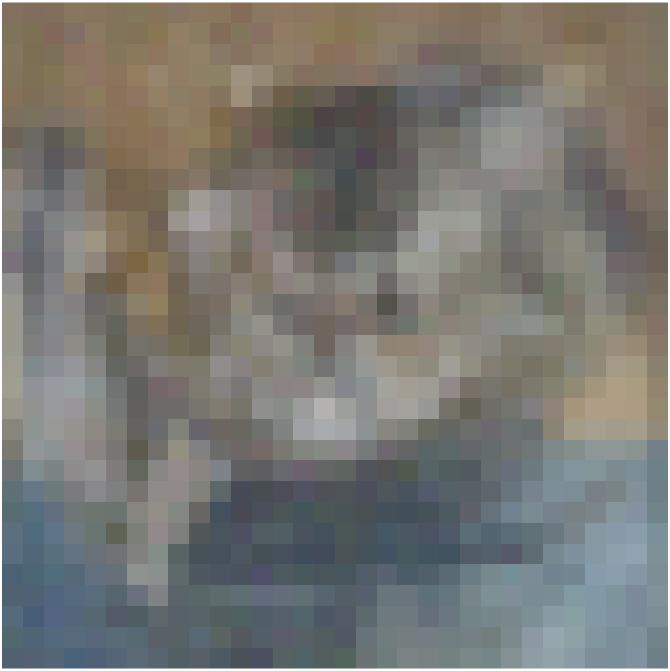}} \:
    \subfigure[2.20]{\label{fig:fs_cif10_te1_path_tr19821_220}%
      \includegraphics[width=0.10\linewidth]{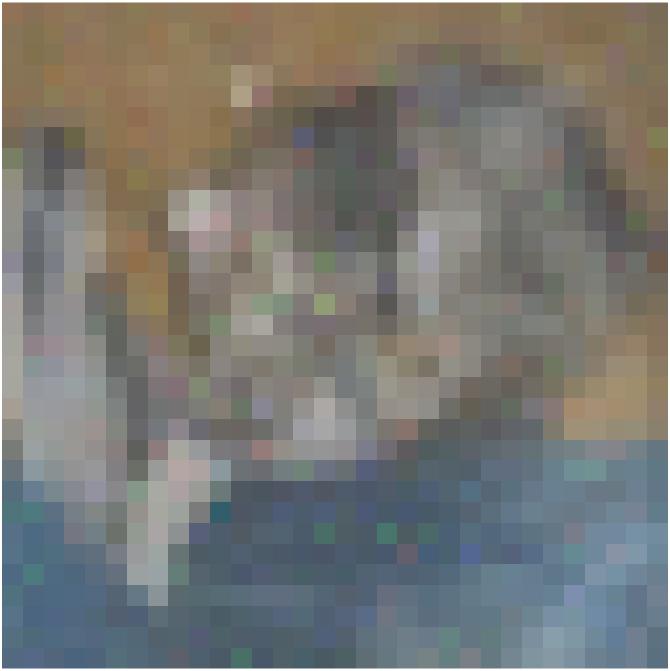}} \:
    \subfigure[2.10]{\label{fig:fs_cif10_te1_path_tr19821_210}%
      \includegraphics[width=0.10\linewidth]{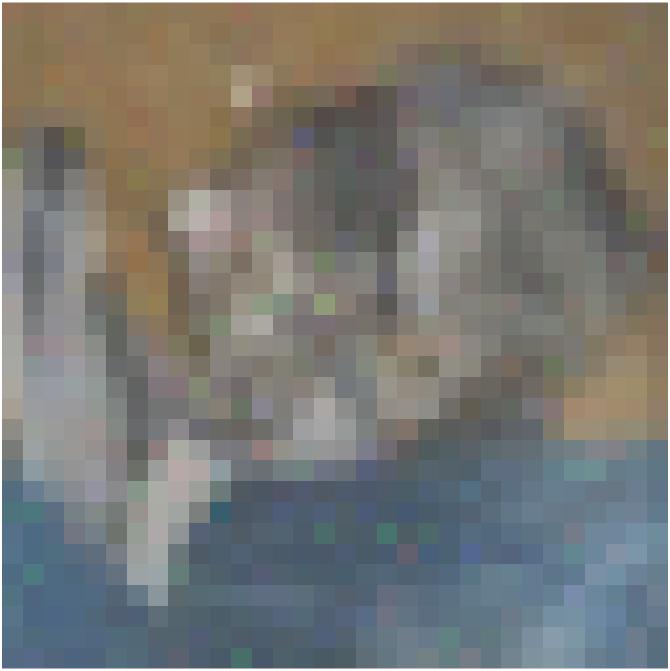}}\\
    \subfigure[2.00]{\label{fig:fs_cif10_te1_path_tr19821_200}%
      \includegraphics[width=0.10\linewidth]{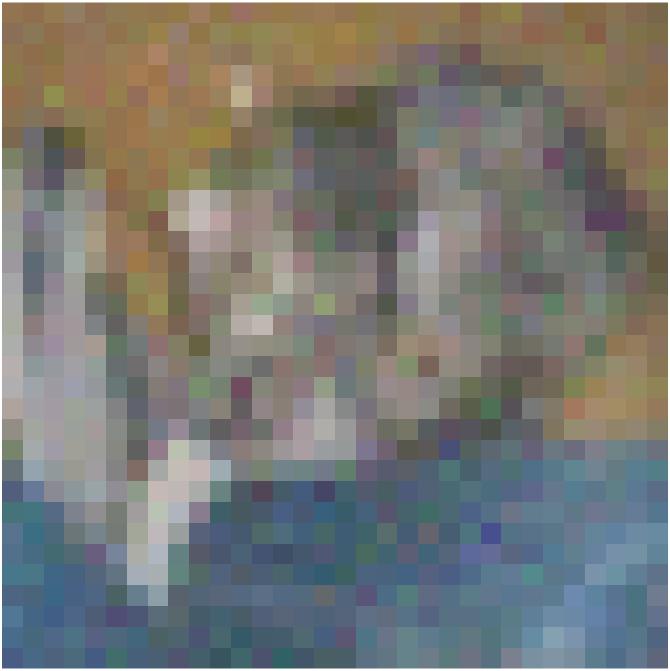}} \:
    \subfigure[1.80]{\label{fig:fs_cif10_te1_path_tr19821_180}%
      \includegraphics[width=0.10\linewidth]{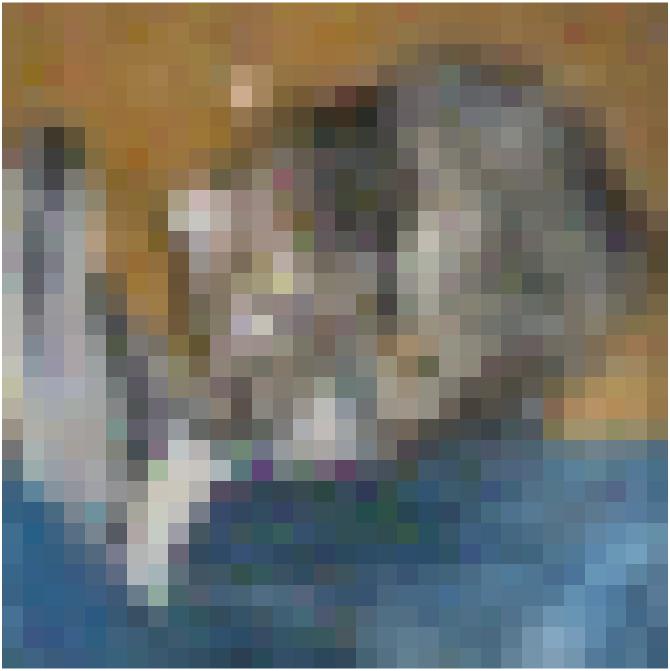}} \:
    \subfigure[1.50]{\label{fig:fs_cif10_te1_path_tr19821_150}%
      \includegraphics[width=0.10\linewidth]{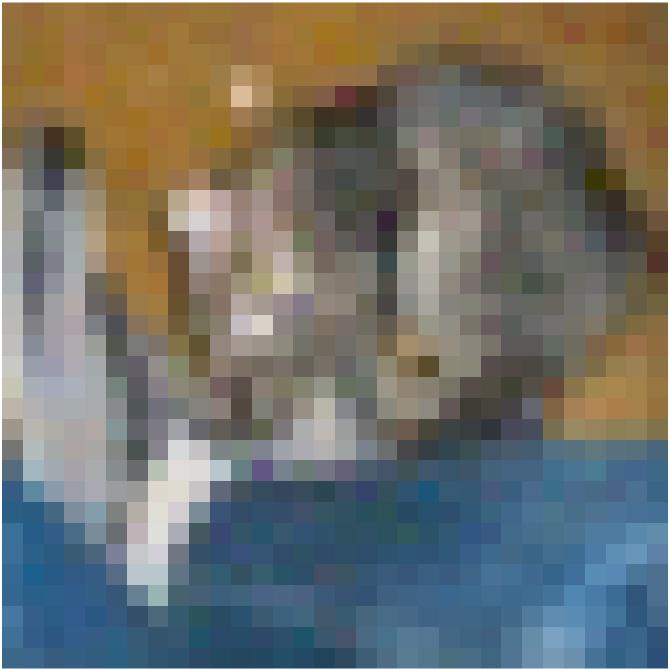}} \:
    \subfigure[1.00]{\label{fig:fs_cif10_te1_path_tr19821_100}%
      \includegraphics[width=0.10\linewidth]{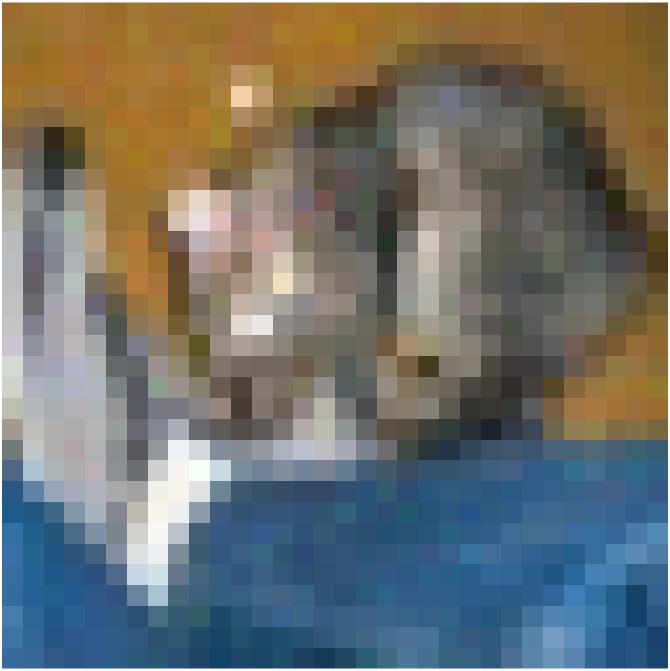}} \:
    \subfigure[0.50]{\label{fig:fs_cif10_te1_path_tr19821_050}%
      \includegraphics[width=0.10\linewidth]{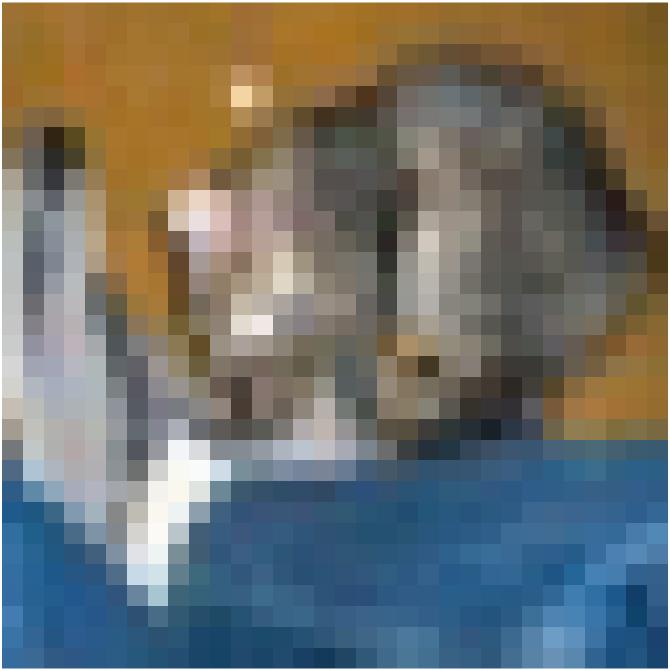}} \:
    \subfigure[0.00]{\label{fig:fs_cif10_te1_path_tr19821_000}%
      \includegraphics[width=0.10\linewidth]{figures/cif10_te1orig.jpg}}
      }
\end{figure}

{\bf Morphing between images.} We now explore the path between two images inside the $\mathcal{B}(x_\phi)\cap \Omega_\phi$. We pick the image shown in Figure~\ref{fig:cif10_tr19821} which is the 19821$^{th}$ sample from the training set. In $\Phi$, this image is distanced 2.546 from $x_\phi$, so it is close to the perimeter of $\mathcal{B}(x_\phi)$. We gradually move between these image in the feature space and find how the path between them maps back to the pixel space. This is done by solving equations~\eqref{eq:ball_c1}-\eqref{eq:ball_obj} while decreasing the value of $r$ from 2.54 to 0. Result is depicted in Figure~\ref{fig:cif10_te1_morph19821}.

% {\bf Morphing between images.} We pick one of the images in training set (19,821$^{th}$ image) and explore the path between that image and our $x_\phi$ which is the first testing sample of dataset. The 19,821$^{th}$ training sample is shown in Figure~\ref{fig:}. The mapping of this image to the feature space falls near the perimeter of $\mathcal{B}(x_\phi)$, barely inside the ball (radis and distance). Therefore, the direct path in feature space between these two images is guaranteed to be classified as Cat by our model $\mathcal{N}$. We now explore this path in feature space and find images in pixel space that would map to that path. Figure~\ref{fig:} shows such images all of which are classified as Cat by the model. Notice how quickly our testing images morphs to the testing image along that path. While 

Note that moving between these images in the feature space leads to a morphing process between them in the pixel space which is more sophisticated than simple image interpolation \citep{lakshman2015image}. Hence, our formulations can be used for image morphing which has practical applications \citep{effland2021image}. Moreover, this transformation is not linear, i.e., change does not occur at a linear rate along the path between the two images. The image in subfigure ($a$) is distanced 2.54 from subfigure~($l$). By the time its distance is 2.2 from ($l$), it appears more similar to ($l$) than ($a$). By the time its distance to ($l$) is 1.80, it looks almost like ($l$) despite its relative closeness to ($a$).

\begin{figure}[h]
  \centering
   \includegraphics[width=0.35\linewidth]{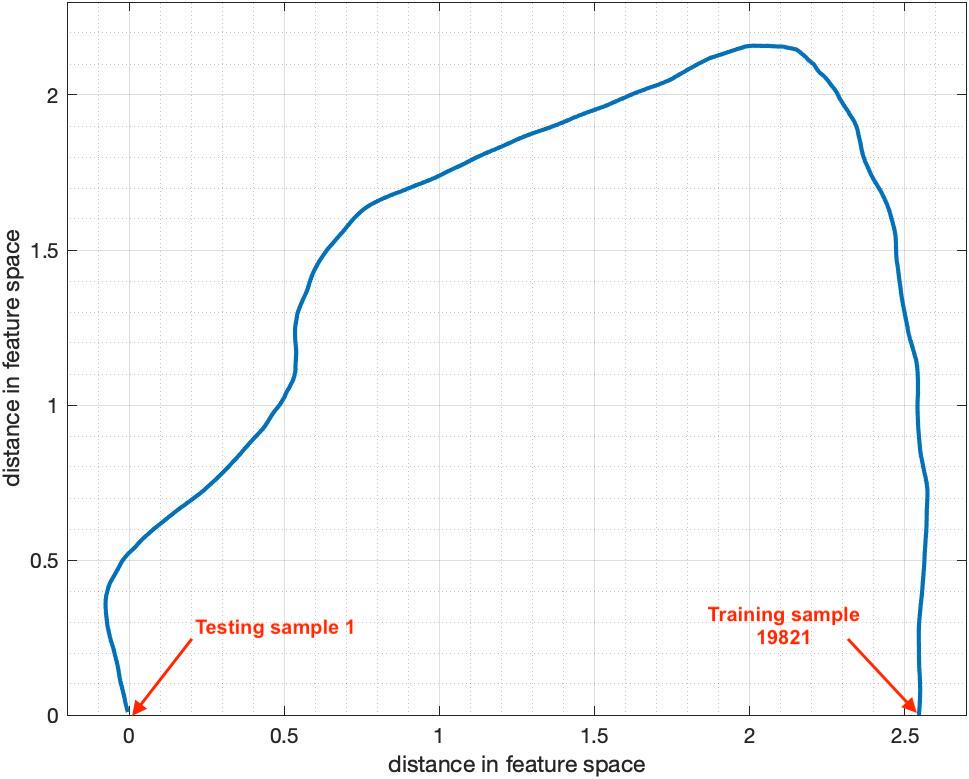}
   \vspace{-.5cm}
   \caption{Direct paths in pixel space map to highly curved paths in the feature space. The blue line shows the direct path between images shown in Figures~\ref{fig:cif10_tr19821} and \ref{fig:fs_cif10_te1_path_tr19821_000}, mapped to the 64-dimensional feature space, then visualized in 2D.}
   \vspace{-.7cm}
  \label{fig:cif10_te1_path_tr19821}
\end{figure}

{\bf Mapping paths from the pixel space to the feature space.} In the previous experiment, we moved between two images in the feature space and saw how they morph in the pixel space. Let us now move between those same images in the pixel space and see how the path between them looks like in the feature space. In the pixel space, we follow a direct path along a line connecting these two images, but as Figure~\ref{fig:cif10_te1_path_tr19821} shows, the resulting path between them in the feature space is far from a direct line. Our feature space, $\Phi$, is 64-dimensional. To draw this path in 2 dimensions, we use the two-point equidistant projection method as explained in Appendix~\ref{sec:appx_mapping}.% The line connecting our two images can be considered one side of a triangle and its length, $d_3$, is fixed. In a 2-dimensional space, we can pick 2 arbitrary points where their distance from each other is $d_3$. Any image, $x_p$, on the path between our two images has a distance to each of them. Let us denote distance of $x_p$ to image 1 with $d_1$ and its distance to image 2 with $d_2$. We have now obtained a triangle with known length for all its sides. Using the {\em law of cosines}, we can derive the internal angles of this triangle, and map $x_p$ to that 2-dimensional space where its distances to image 1 and image 2 are $d_1$ and $d_2$, respectively. When $d_1+d_2 = d_3$, the path between image 1 and image 2 follows a direct line, i.e., it is linear. However, when $d_1+d_2 > d_3$, the path between our images will be curved. Appendix~\ref{sec:appx_mapping} explains this in more detail.

% Figure~\ref{fig:cif10_te1_path_tr19821} shows the mapping of the path between our two images in Figure~\ref{fig:cif10_te1_morph19821}. In the pixel space, we have followed a direct path between these two images, however, their projection to the feature space is far from a direct path.

\subsection{Larger Trends in the CIFAR-10 dataset} \label{sec:results_cif10_trends}

We extend this analysis to the entire dataset to see the larger trends persistent for most images. 

\begin{figure}[h]
\floatconts
  {fig:fs_cif10_fliphullmargins}
  {    \vspace{-.7cm}
\caption{Distribution of distance in the feature space {\bf (a)} to closest flip point, {\bf (b)} to convex hull of training set. For most samples, decision boundaries are much further away than the convex hull of training set. In the pixel space, however, this relationship is reversed.} \vspace{-.5cm}}
  {%
    \subfigure[]{\label{fig:fs_cif10_flipmargins}%
      \includegraphics[width=0.4\linewidth]{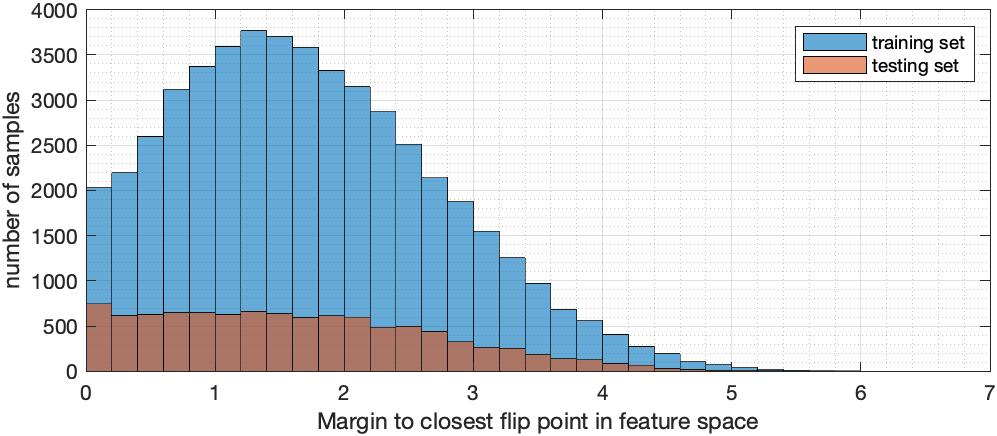}} \quad
    \subfigure[]{\label{fig:fs_cif10_hullmargins}%
      \includegraphics[width=0.4\linewidth]{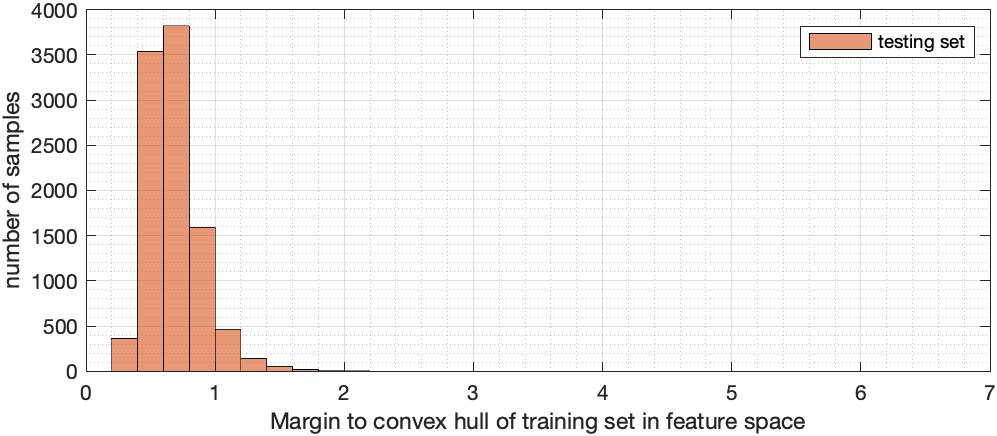}}
      }
\end{figure}

{\bf Geometric arrangements in the feature space.} Figures~\ref{fig:fs_cif10_flipmargins}-\ref{fig:fs_cif10_hullmargins} show that for most testing samples, distance to the closest decision boundaries is larger than the distance to convex hull of training set. This difference has broad implications. For example, when we project testing samples to the convex hull of training set in the pixel space, testing accuracy of the model drops from above 90\% to 33\% on those projected images. However, when we project the testing samples to the convex hull of training set in the feature space, the accuracy does not change at all, meaning that in the feature space, model has not defined any decision boundaries separating testing samples from their projections to the $\mathcal{H}^{tr}_\phi$.

% \begin{figure}[h]
%     \centering
%     \includegraphics[width=.45\textwidth]{figures/cif10_dist_fs_flip.jpg}
%     \caption{Distribution of distance to closest flip point in the feature space.}
%     \label{fig:fs_cif10_flipmargins}
% \end{figure}

% \begin{figure}[h]
%     \centering
%     \includegraphics[width=.45\textwidth]{figures/cif10_dist_fs_hull.jpg}
%     \caption{Distribution of distance to the convex hull of the training set in the feature space for samples of CIFAR-10 dataset. Comparing these distances to Figure~\ref{fig:fs_cif10_flipmargins} reveals that for most samples, decision boundaries are much further away than the convex hull of the training set. In the pixel space, however, this relationship is reversed.}% Closeness to decision boundaries makes models more vulnerable to adversarial examples.}
%     \label{fig:fs_cif10_hullmargins}
% \end{figure}

\begin{figure}[h]
\floatconts
  {fig:cif10_te732}
  {    \vspace{-.5cm}
\caption{{\bf (a)} Testing sample \#732, {\bf (b)} Mapping of (a) to $\Phi$, {\bf (c)} Modified version of (a) to remove its ambiguity, {\bf (d)}~Mapping of (c) to $\Phi$.} \vspace{-.7cm}}
  {%
    \subfigure[]{\label{fig:cif10_te732_orig}%
      \includegraphics[width=0.10\linewidth]{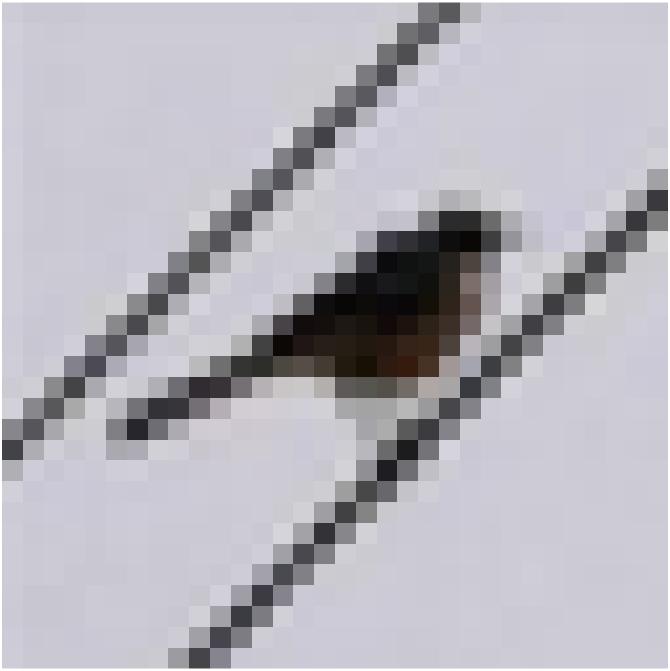}}
    \subfigure[]{\label{fig:cif10_te732_fs}%
      \includegraphics[width=0.11\linewidth]{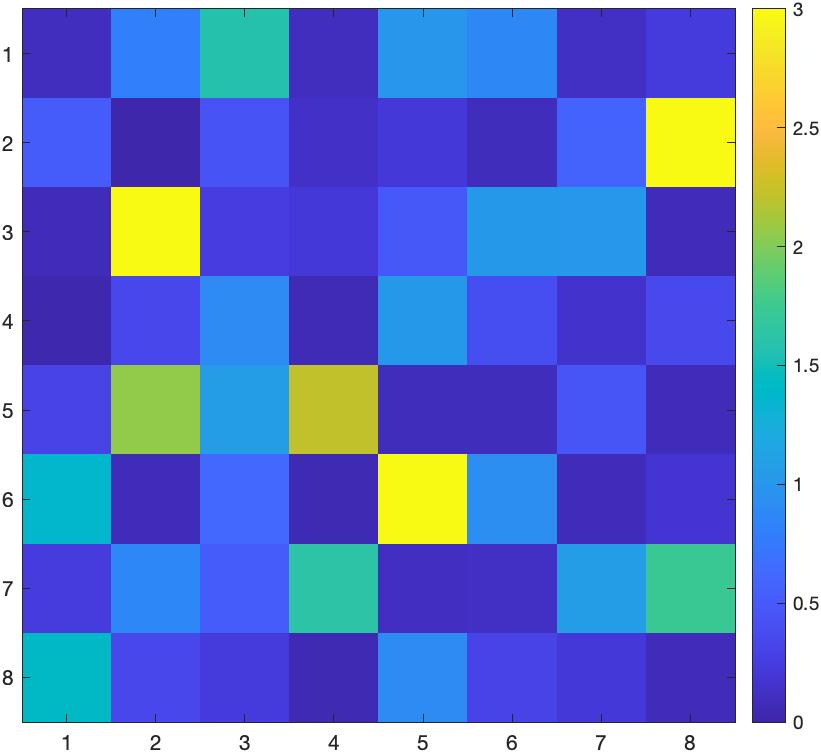}} \quad
    \subfigure[]{\label{fig:cif10_te732_modif}%
      \includegraphics[width=0.10\linewidth]{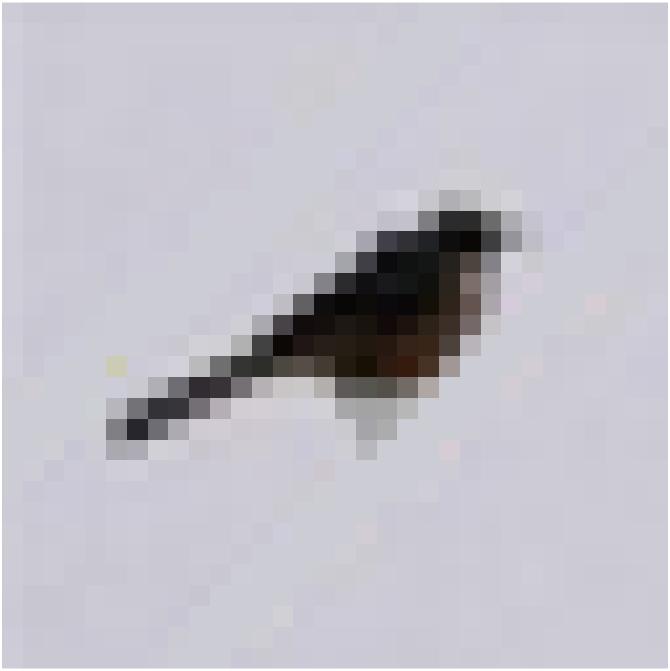}}    
    \subfigure[]{\label{fig:cif10_te732_modif_fs}%
      \includegraphics[width=0.11\linewidth]{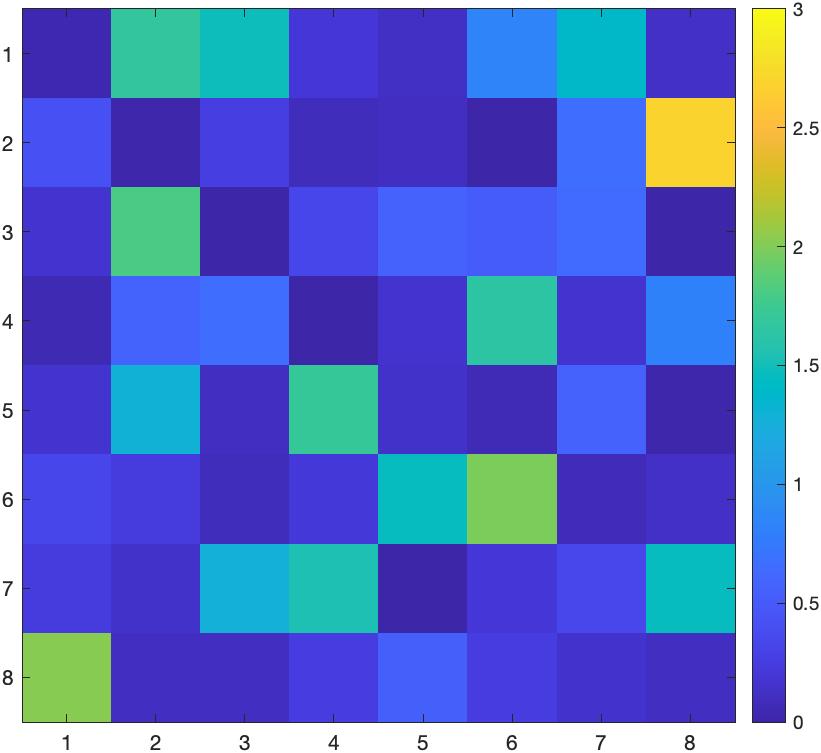}}
      }
\end{figure}

{\bf Detecting ambiguous images.} In feature space, convex hull of the training set is closer than the decision boundaries for 78.3\% of testing samples. Let us see what is different about the remaining 21.7\% of images. Testing sample \#732, shown in Figure~\ref{fig:cif10_te732_orig}, is distanced 0.3745 from the closest decision boundary in $\Phi$ while its distance to the $\mathcal{H}^{tr}_\phi$ is 2.143. This is clearly an ambiguous image from the model's perspective, because in the feature space, this image is very close to model's decision boundaries, yet very far from the training set.

% \begin{figure}[h]
%      \centering
%      \begin{subfigure}[b]{0.15\linewidth}
%          \centering
%          \includegraphics[width=\linewidth]{figures/cif10_te732.jpg}
%          \caption{}
%          \label{fig:cif10_te732_orig}
%      \end{subfigure}
%      \begin{subfigure}[b]{0.164\linewidth}
%          \centering
%          \includegraphics[width=\linewidth]{figures/cif10_te732_fs.jpg}
%          \caption{}
%          \label{fig:cif10_te732_fs}
%      \end{subfigure}
%      \begin{subfigure}[b]{0.15\linewidth}
%          \centering
%          \includegraphics[width=\linewidth]{figures/cif10_te732_modif.jpg}
%          \caption{}
%          \label{fig:cif10_te732_modif}
%      \end{subfigure}
%      \begin{subfigure}[b]{0.164\linewidth}
%          \centering
%          \includegraphics[width=\linewidth]{figures/cif10_te732_modif_fs.jpg}
%          \caption{}
%          \label{fig:cif10_te732_modif_fs}
%      \end{subfigure}
%      \caption{{\bf (a)} Testing sample \#732, {\bf (b)} Mapping of (a) to $\Phi$, {\bf (c)} Modified version of (a) to remove its ambiguity, {\bf (d)}~Mapping of (c) to $\Phi$.}
%     \label{fig:cif10_te732}
% \end{figure}

% \vspace{-.5cm}

From a human's perspective, as opposed to the model's, ambiguity may be perceived differently because, a human typically have seen many instances of birds and alike, in different settings/contexts and against various backgrounds. However, the model trained on the CIFAR-10 training set has only seen 5,000 bird images, and the testing image \#732 is not similar to any training image regarding the parallel wires below and above the bird. Therefore, this testing image can be considered ambiguous.

Let us now try to remove the ambiguity by eliminating the parallel wires as shown in Figure~\ref{fig:cif10_te732_modif}. Mapping of this modified image to the feature space is drastically different than the mapping of original image. In fact, in $\Phi$, these two images are 5.21 apart which is considerable compared to those distances we previously reported for other images (e.g., in Figures~\ref{fig:cif10_te1_path_tr19821} and~\ref{fig:fs_cif10_flipmargins}). The modified image is only distanced 0.605 from the $\mathcal{H}^{tr}_\phi$ while its distance to the closest flip point has drastically increased to $1.225$ (the flip point to the Airplane class). From the model's perspective, our modification has removed the ambiguity from the image because now, in the feature space, the image is much closer to the convex hull of training set and it has also moved away from the decision boundaries.

\begin{figure}[h]
  \centering
   \includegraphics[width=0.5\linewidth]{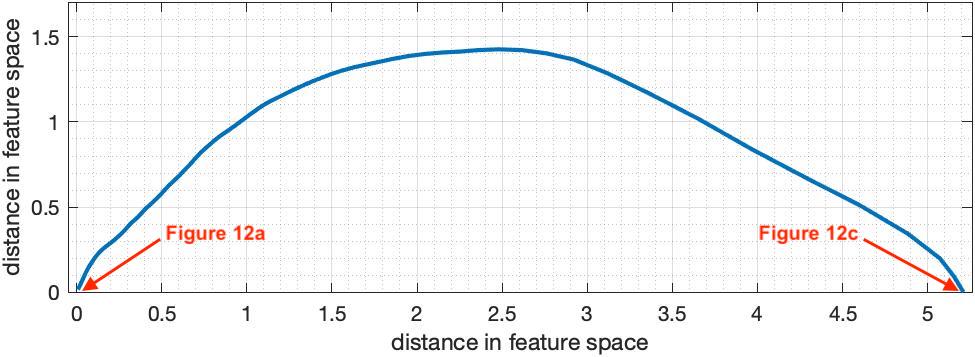}
   \vspace{-.5cm}
   \caption{Visualization of direct path between images shown in Figures~\ref{fig:cif10_te732_orig} and \ref{fig:cif10_te732_modif}.} 
   \vspace{-.2cm}
  \label{fig:cif10_te732_path_modified}
\end{figure}

Figure~\ref{fig:cif10_te732_path_modified} shows the visualized path in $\Phi$ between the testing image \#732 and its unambiguous counterpart. As we can see, the path between these images is nonlinear even though moving between them is merely, gradual removal of the wires. But note that non-linearity of the path is more moderate in comparison to the path in Figure~\ref{fig:cif10_te1_path_tr19821}.

%It happens that this image is also considered ambiguous for humans from the cognitive science perspective as reported via an empirical study by \citep{}. In that cognitive science study, humans were presented with certain number of training samples of CIFAR-10 and then, were asked to classify several testing images of CIFAR-10 in a time frame. Ambiguity for each image was characterized based on the correctness of human classifications and the time it took for them to classify the image.

{\bf Formalizing an ambiguity indicator.} This leads us to consider the difference between the distance to closest flip point in $\Phi$ and the distance to $\mathcal{H}^{tr}_\phi$ as a relative indicator for ambiguity
\begin{equation} \label{eq:dist_ambig}
    d_\phi^{f-h} = d^{f,min}_\phi (x_\phi) - d^{h}_\phi,
\end{equation}
drawing from the distances previously defined by equations~\eqref{eq:fb_dmin} and~\eqref{eq:dist_hull}. Figure~\ref{fig:cif10_ambig} shows images with extreme values of $d_\phi^{f-h}$.

\begin{figure}[h]
\floatconts
  {fig:cif10_ambig}
  {    \vspace{-.5cm}
\caption{Images with the largest values of $d_\phi^{f-h}$ which we consider to be an ambiguity indicator. Images in (a) are close to model's decision boundaries and far from the convex hull of the training set in the feature space. Images in (b) are far from model's decision boundaries yet very close to the convex hull of training set in $\Phi$.}\vspace{-.5cm}}
  {%
    \subfigure[Largest negative values of $d_\phi^{f-h}$ (most ambiguous)]{\label{fig:cif10_ambig_most}%
      \includegraphics[width=0.10\linewidth]{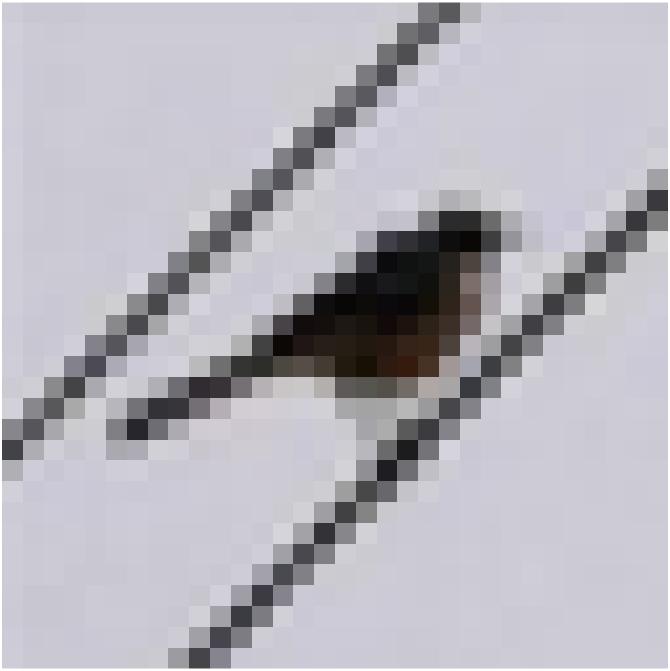}
      \includegraphics[width=0.10\linewidth]{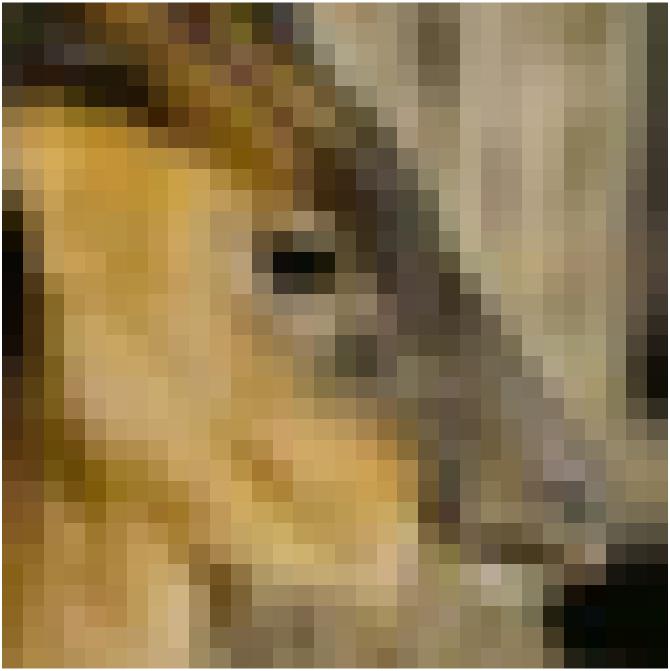}
      \includegraphics[width=0.10\linewidth]{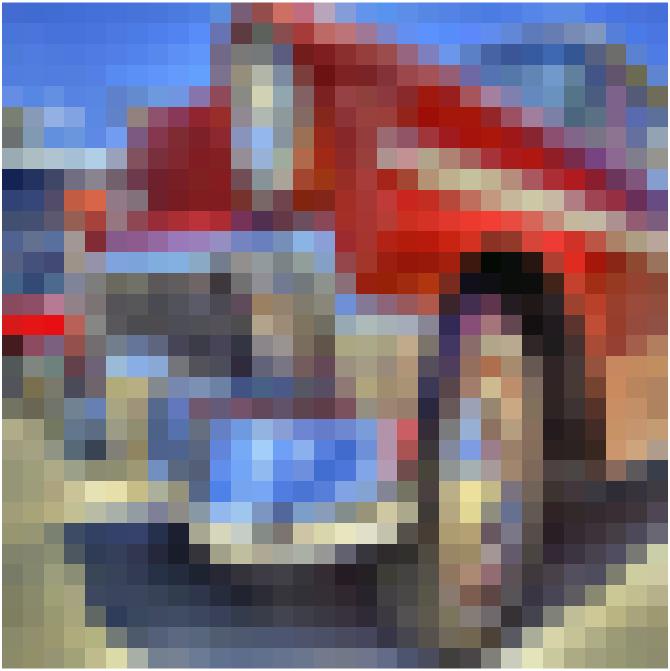}
      \includegraphics[width=0.10\linewidth]{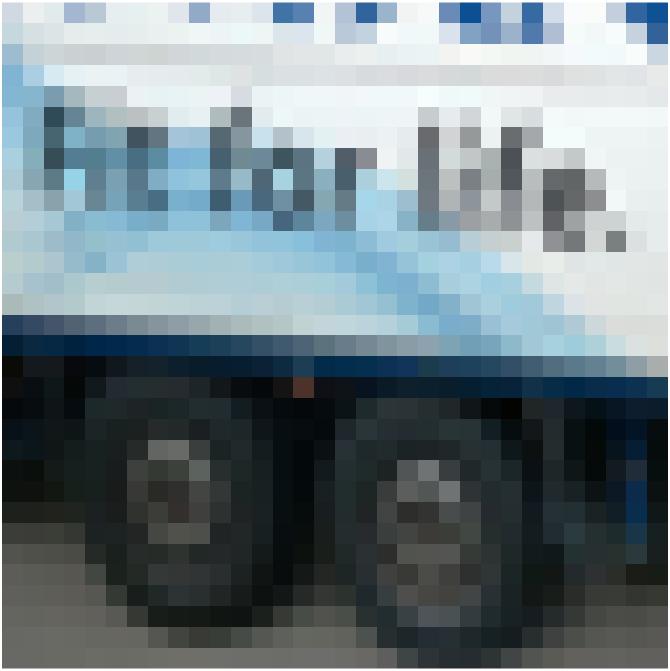}
      \includegraphics[width=0.10\linewidth]{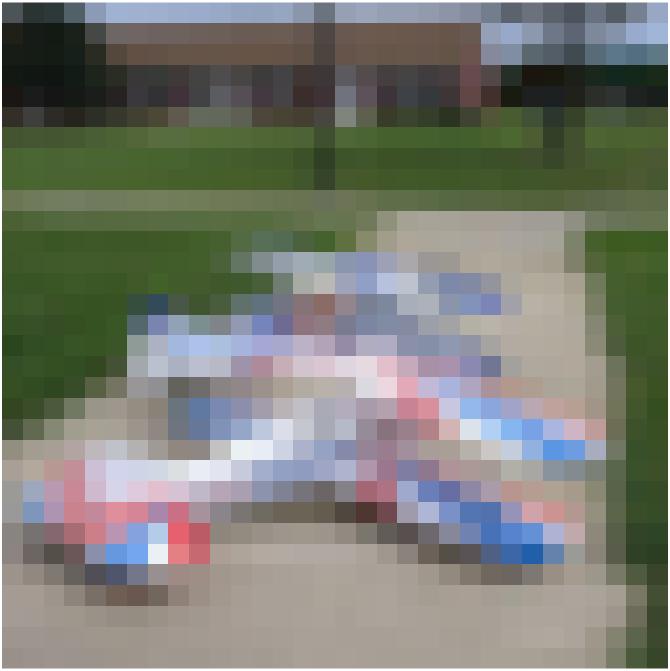}
      \includegraphics[width=0.10\linewidth]{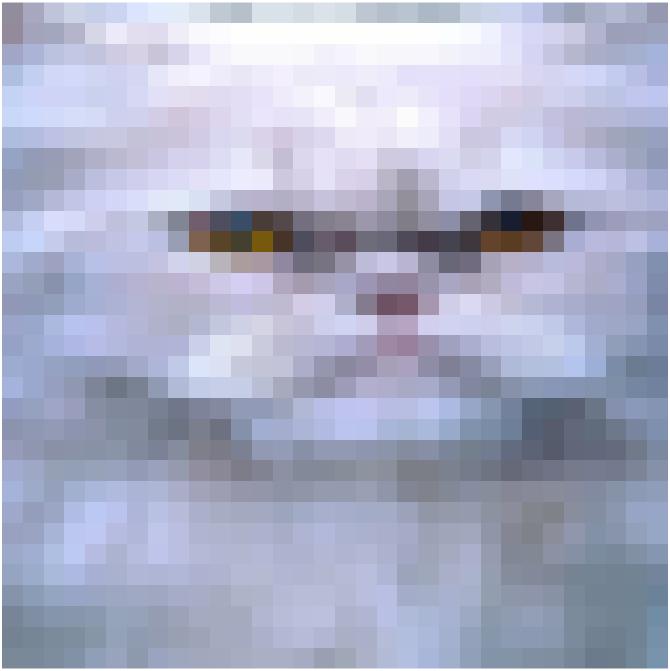}
      } \;
    \subfigure[Largest positive values of $d_\phi^{f-h}$ (least ambiguous)]{\label{fig:cif10_ambig_least}%
      \includegraphics[width=0.10\linewidth]{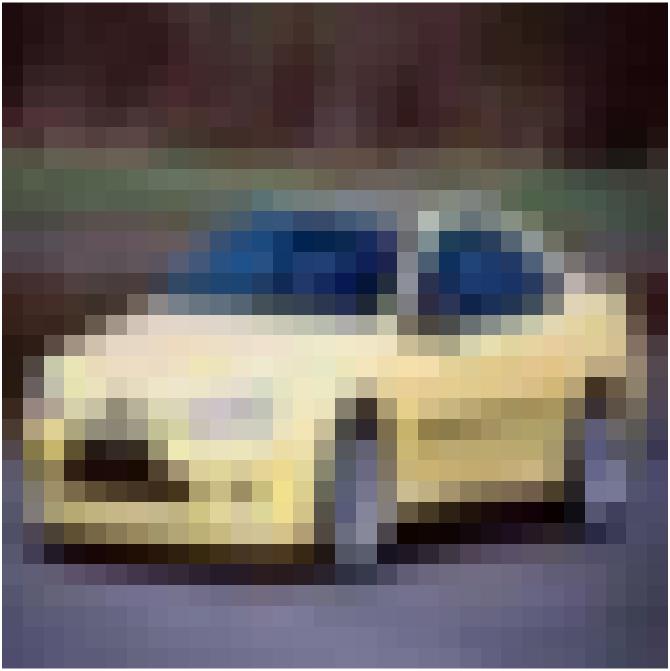}
      \includegraphics[width=0.10\linewidth]{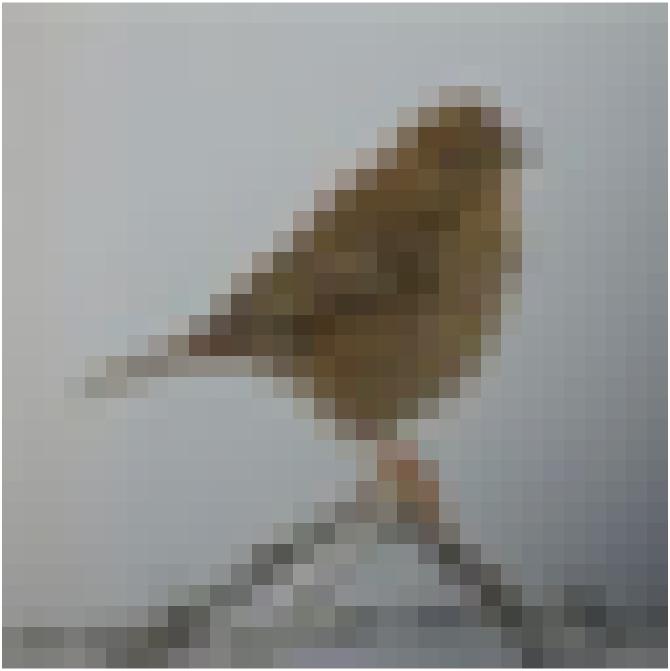}
      \includegraphics[width=0.10\linewidth]{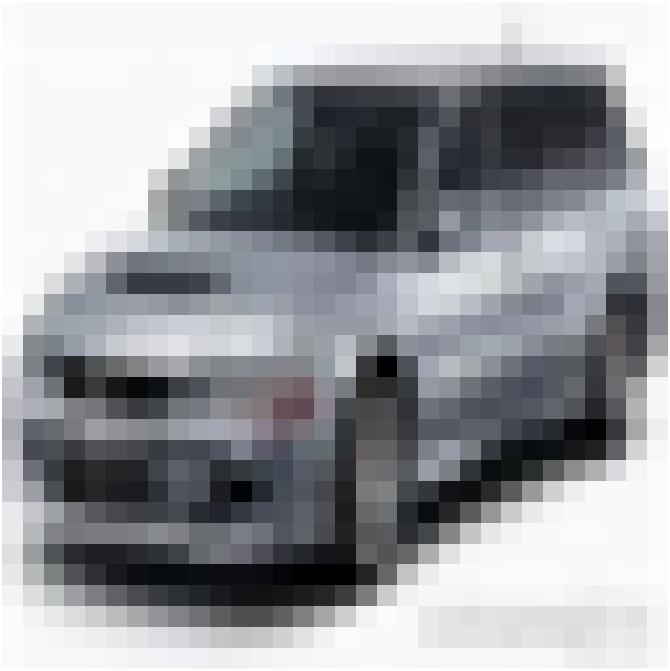}
      \includegraphics[width=0.10\linewidth]{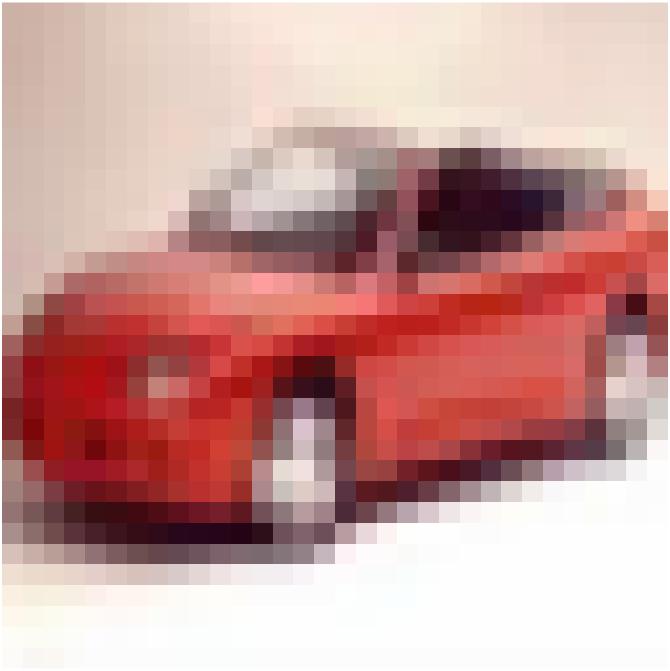}
      \includegraphics[width=0.10\linewidth]{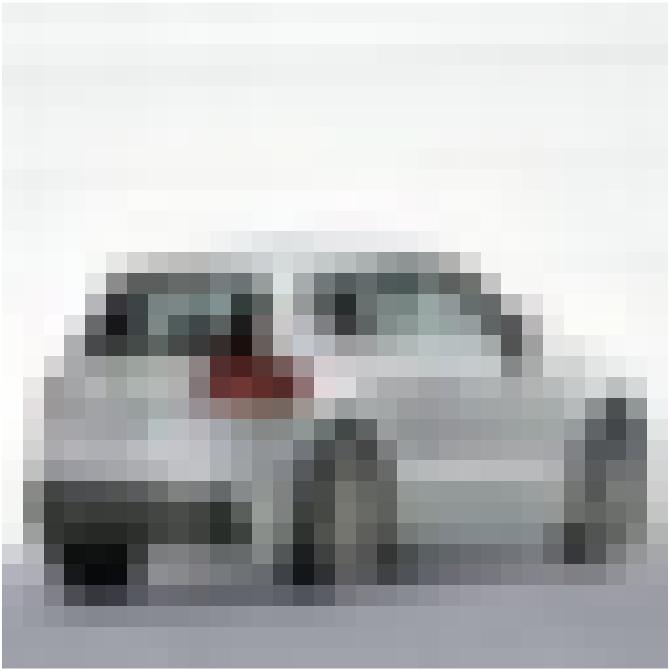}
      \includegraphics[width=0.10\linewidth]{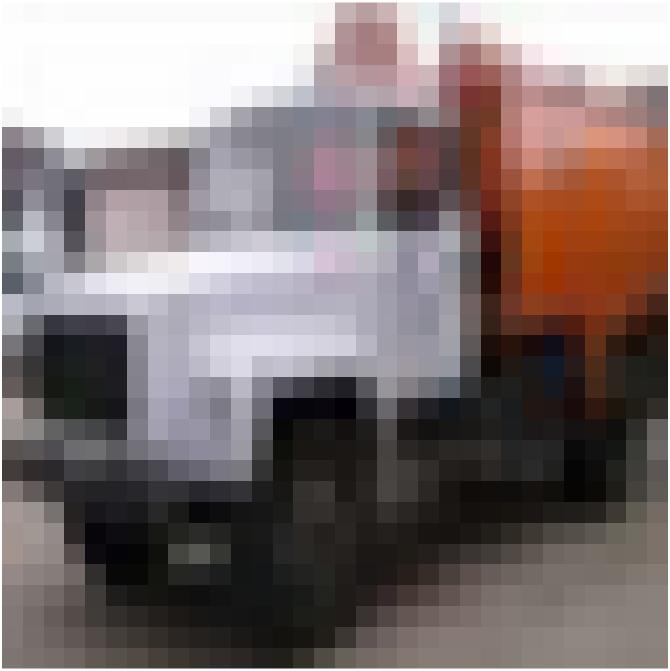}
      }}
\end{figure}

% \vspace{-.5cm}

This notion of ambiguity takes into account closeness to decision boundaries in the feature space learned by the model and contrasts it with the farness from $\mathcal{H}^{tr}_\phi$. When an image falls close to a decision boundary in a feature space, the model may be unsure about the classification, because the image may easily cross the close-by decision boundary and fall into the partition for a different class. Regarding the $\mathcal{H}^{tr}_\phi$, when an image falls close to $\mathcal{H}^{tr}_{\phi}$, it means that the model has a close point of reference to it in the training set, and therefore, can be more confident in the correctness of classification. By this logic and from the trained model's perspective, all images in Figure~\ref{fig:cif10_ambig_most} can be considered ambiguous while all images in Figure~\ref{fig:cif10_ambig_least} can be considered unambiguous.

We note that most ambiguous images we report are previously reported to be ambiguous for humans from the cognitive science perspective via empirical studies by \citet{peterson2019human,battleday2020capturing}. In those cognitive science studies, humans were presented with a certain number of training samples of CIFAR-10, and then, were asked to classify testing images of CIFAR-10 in a certain time frame. Ambiguity of images was characterized based on the correctness of human classifications and the time it took for humans to classify them. This can be the subject of further study from the perspective of cognitive science and psychology as well as machine learning. Identifying ambiguous images are also useful in practice.

{\bf Detecting adversarial examples.} Geometric arrangements in the feature space have implications for detecting adversarial images. \emph{Our suggested rule of thumb is that any testing image very close to a decision boundary is likely to be an adversarial input.} %, especially if it is close to or falls inside the $\mathcal{H}^{tr}_\phi$.} 
For testing samples in this dataset, we see that using a simple threshold, we can detect all adversarial inputs.

\begin{figure}[h]
\floatconts
  {fig:cif10_te2}
  {    \vspace{-.5cm}
\caption{{\bf (a)} Testing sample \#2, {\bf (b)} Adversarial version of it classified as Airplane.\vspace{-.5cm}}}
  {%
    \subfigure[]{\label{fig:cif10_te2_orig}%
      \includegraphics[width=0.12\linewidth]{figures/cif10_te2.jpg}}
 \quad \qquad \qquad \qquad \qquad
    \subfigure[]{\label{fig:cif10_te2_adv}%
      \includegraphics[width=0.12\linewidth]{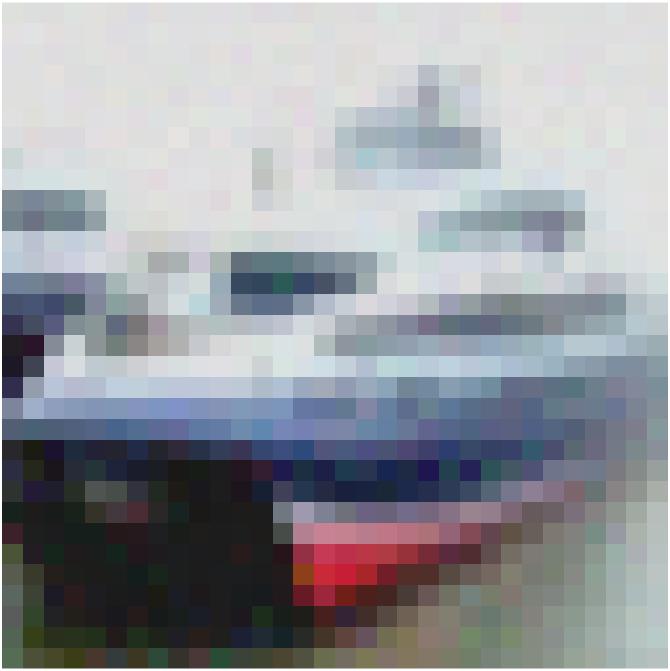}}
      }
\end{figure}

Consider, for example, the image in Figure~\ref{fig:cif10_te2_orig} and its adversarial counterpart in Figure~\ref{fig:cif10_te2_adv} classified as Airplane. The original image is distanced 2.494 from the closest decision boundary while its distance to $\mathcal{H}^{tr}_\phi$ is 0.845. On the other hand, its adversarial version is distanced 0.0001 from the closest decision boundary in $\Phi$ while its distance to $\mathcal{H}^{tr}_\phi$ is 0.640. Extreme closeness of this sample to the decision boundary in $\Phi$ is a clear indication of its adversary nature.% At the same time, the distance to $\mathcal{H}^{tr}_\phi$ can be used as a frame of reference about distances in $\Phi$.

% \begin{figure}[h]
%      \centering
%      \begin{subfigure}[b]{0.3\linewidth}
%          \centering
%          \includegraphics[width=.5\linewidth]{figures/cif10_te2.jpg}
%          \caption{}
%          \label{fig:cif10_te2_orig}
%      \end{subfigure}
%     %  \quad \quad \qquad \qquad \qquad \qquad
%      \begin{subfigure}[b]{0.3\linewidth}
%          \centering
%          \includegraphics[width=.5\linewidth]{figures/cif10_te2_adv.jpg}
%          \caption{}
%          \label{fig:cif10_te2_adv}
%      \end{subfigure}
%     %  \begin{subfigure}[b]{0.45\textwidth}
%     %      \centering
%     %      \includegraphics[width=\textwidth]{figures/cif10_path_fs_te3_adv.jpg}
%     %      \caption{}
%     %      \label{fig:cif10_te3_adv_path}
%     %  \end{subfigure}
%     \caption{{\bf (a)} Testing sample \#2, {\bf (b)} Adversarial version of it classified as Airplane.}
%     \label{fig:cif10_te2}
% \end{figure}

% \vspace{-.5cm}

In the pixel space, genuine images are often very close to decision boundaries, so closeness to decision boundaries is not a good measure to distinguish adversarial inputs from genuine ones. In the feature space, however, decision boundaries are relatively far from the images, especially in comparison to the $\mathcal{H}^{tr}_\phi$. In other words, an image very close to the decision boundaries of feature space is unusual, and this closeness can be used as an indicator.

For 100\% of testing samples, their adversarial version is closer to the decision boundaries of the feature space compared to the $\mathcal{H}^{tr}_\phi$. Their margin to decision boundaries is also closer than the margin to decision boundaries for all training/testing samples. In other words, adversarial methods move the testing samples, recognizably, very close to the decision boundaries of the feature space, by any of these measures of comparison. 

Moreover, we see that standard adversarial methods such as DeepFool move the samples towards the $\mathcal{H}^{tr}_\phi$. See Appendix~\ref{sec:appx_adv} for further discussion.% At the same time, adversarial methods move the images closer to $\mathcal{H}^{tr}_\phi$ as well. 

{\bf Union of learned regions.} Earlier, we mentioned that a classification model is a function defined by its decision boundaries. A model learns from the contents of training images and their labels. Via this process, model partitions the domain (in pixel space and in feature space) by defining certain decision boundaries. We defined the ball around each image that borders with the closest decision boundary. Such ball can be viewed as a region known to the model and guaranteed to have a certain classification. We reported earlier that for the first testing sample of dataset, the radius of that ball was quite large in the feature space such that it contained hundreds of training and testing samples, having a considerable overlap with the convex hull of training set. This trend holds for many other images in the dataset. In fact, for 49.9\% of training samples, their corresponding $\mathcal{B}(x_\phi)$ contains at least another training or testing sample. Similarly, for 47\% of testing samples, their $\mathcal{B}(x_\phi)$ contains other training/testing samples. On average, each $\mathcal{B}(x_\phi)$ contains about 297 other training and testing samples. The largest number of samples contained in a $\mathcal{B}(x_\phi)$ is 4,791.

Therefore, the learned regions, defined by $\mathcal{B}(x_\phi)$ around each image, have significant overlaps, and we can study the {\em union of learned regions} defined by $$\bigcup_{i=1}^{n} \mathcal{B}(x_i),$$ for all the $n$ samples in a training set.

Overall, more than 68\% of testing samples are contained in the union of learned regions for the training set. These samples could be considered most familiar samples for the model as they fall into familiar regions in the feature space relating closely to training samples. This concept may also be useful for detecting out-of-distribution images, and images with low-confidence in their classification.

{\bf How images are supported by the convex hull of training set.} Using equation~\eqref{eq:project_hull}, we project each testing image to the convex hull of training set. We perform this both in the pixel space and in the feature space. Projection of each image to the convex hull is a point defined as a convex combination of certain support images in the training set. We see that in the feature space, 78\% of support images have the same label as the testing image that they are supporting while this percentage is only 27\% in the pixel space. This shows that in the feature space, a testing image of a given class, let us say Automobile, is supported mostly by training images of Automobile class, whereas in the pixel space, a testing image from the Automobile class may be supported by training images of many other classes. This is another evidence that geometric arrangement of images in the feature space is more sensible and meaningful from the classification perspective.

\vspace{-.2cm}

\section{Conclusions}

In this work, we presented a set of formulations that can answer questions about the inner workings of feature space learned by trained neural networks. Our formulations incorporate any trained model as a function, and find images in the pixel space that map to particular points and regions of interest in the feature space. This enabled us to provide many novel insights about image classification functions, the features that they learn from images, and their adversarial vulnerabilities. Although our formulations about the pixel space are generally hard to solve, we were able to solve them with a homotopy algorithm. The feature space, on the other hand, is Lipschitz continuous with a known constant which enable us to study it with clarity. We identify certain regions around each image guaranteed to have the same classification as the image. We then investigated these regions with respect to training samples and decision boundaries of the model. Notably, we observed that geometric arrangements of decision boundaries are considerably different in the feature space in relation to training and testing samples, providing a way to identify ambiguous and adversarial images. These geometric arrangements are very different than the ones reported in the literature for the pixel space. Moreover, a new direction of research would be to study adversarial examples that remain far from the decision boundaries and at the same time, maintain a distance from the convex hull of training set in feature space. Finally, these insights may inform us about the functional task of models and the extent in which they extrapolate to classify unseen images.

\bibliography{refs}

\clearpage

\appendix

\section{Two-point equidistant projection} \label{sec:appx_mapping}

This is a standard type of projection which falls under the category of azimuthal projections \citep{close1921note} with wide applications in fields such as cartography \citep{snyder1997flattening} and mathematics \citep{li2021dipole}. In this projection, there are two control points, and projection is performed such that distance of all points from the two control points is preserved. We use this type of projection to visualize (in 2D) and investigate paths between pairs of images in the feature space while using each image as a control point.

%In our experiments, the 2D projection is performed from the 64-dimensional feature space, and not directly from the pixel space.
%The feature space we investigated in this paper is 64 dimensional which is relatively dimensional compared to the few thousands dimensions of pixel space. Yet, it may be helpful to use a 2D space to visualize the paths between 

Consider an $f$-dimensional space, $\Phi$, where we have two points of interest: $A$ and $B$. We would like to explore the path between these two points. The Euclidean distance between $A$ and $B$ is a scalar denoted by $$d_{AB} = d_{BA} = \|B-A\|_2,$$ which is the length of the line $AB$. 

Let us now pick two arbitrary points $a$ and $b$ in a 2D space, $\mathcal{V}$, that are distanced $d_{AB}$ from each other. For simplicity, we can assume that $a$ is located at coordinates $(0,0)$ of $\mathcal{V}$ and $b$ is located at coordinates $(d_{AB},0)$.

\begin{figure}[h]
    \centering
    \includegraphics[width=.5\textwidth]{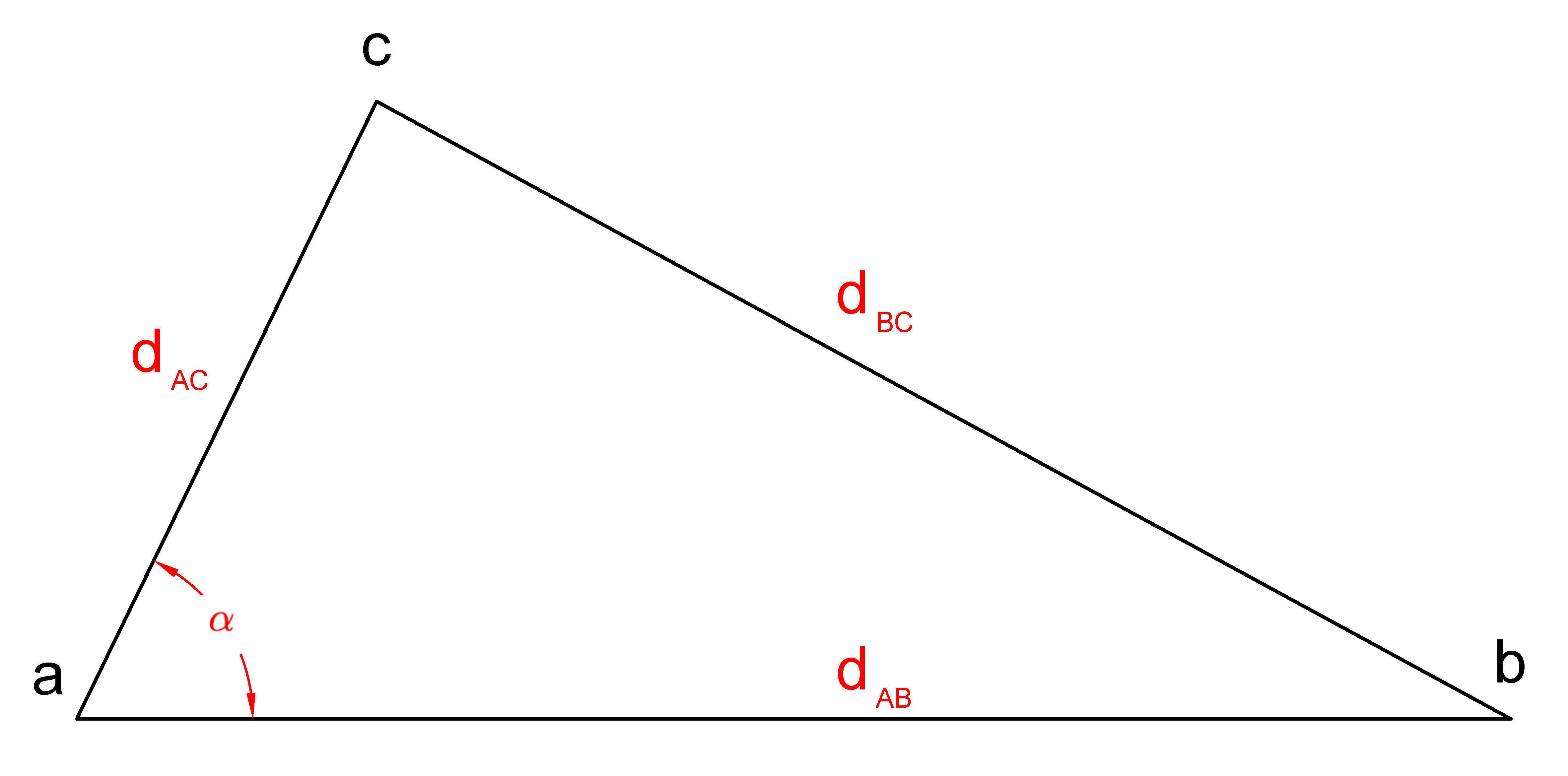}
    \caption{Projecting 3 points from a high dimensional space to a 2D space $\mathcal{V}$ for visualization purposes. We use the two-point equidistant projection method.}
    \label{fig:2d_projection}
\end{figure}

% \textcolor{red}{$\alpha$}

Any other point $C$ in $\Phi$ can be viewed specifically in relation to $A$ and $B$. By the laws of Euclidean space, we have $d_{AB} \leq d_{AC} + d_{BC}$. Now, we would like to find a point $c$ in $\mathcal{V}$ such that its distances to $a$ and $b$ are $d_{AC}$ and $d_{BC}$, respectively. We can find such point using the law of cosines
% \vspace{-.2cm}
\begin{equation} \label{eq:2d_prj}
    \cos (\alpha) = (d_{AC}^2 + d_{AB}^2 - d_{BC}^2) / (2 d_{AC} d_{AB})
\end{equation}
where $\alpha$ is the angle of triangle facing the edge $bc$.

Coordinates of $c:(c_1,c_2)$ can then be obtained via
\begin{equation} \label{eq:2dPrj_c1}
    c_1 = d_{AC} \cos(\alpha),
\end{equation}
% \vspace{-.2cm}
\begin{equation} \label{eq:2dPrj_c2}
    c_2 = d_{AC} \sin(\alpha).
\end{equation}
This way, we can map any point in $\Phi$ to $\mathcal{V}$ while maintaining its distances to $a$ and $b$.

{\bf Comparison of direct and indirect paths.} When $d_{AB} = d_{AC} + d_{BC}$, it implies that in the $f$ dimensional space of $\Phi$, $C$ lies exactly on the direct line connecting $A$ and $B$. Similarly, in $\mathcal{V}$, $c$ will lie exactly on the line connecting $a$ and $b$ because $\alpha$ becomes zero. It follows that for a path in $\Phi$ that entirely lies on the line $AB$, its projection to $\mathcal{V}$ will also entirely lie on the line $ab$.

However, when we have $d_{AB} < d_{AC} + d_{BC}$ for a point $C$, that point is strictly away from the line $AB$, and its projection to $\mathcal{V}$ will also be strictly away from the line $ab$. In fact, as equation~\eqref{eq:2dPrj_c2} shows, larger distance of $c$ from $ab$ will indicate larger deviation of $C$ from the line $AB$ and vice versa because $\alpha$ will be strictly positive, and for a given $A$ and $B$, $\alpha$ will monotonically increase as $C$ deviates farther from $AB$. This rule about points also extends to any paths between $A$ and $B$. When a path between $A$ and $B$ deviates significantly from the line $AB$, its projection to $\mathcal{V}$ will also significantly deviate from the line $ab$.%, as we can see in examples shown in Figure~\ref{fig:}.

% figure

We use this method of projection in Section~\ref{sec:results} and Figures~\ref{fig:cif10_te1_path_tr19821} and \ref{fig:cif10_te732_path_modified} to visualize the paths between various images and see how such paths deviate from a direct line.

\section{Implications for detecting adversarial attacks} \label{sec:appx_adv}

{\bf Closeness to the convex hull of training set.} We reported earlier that all testing samples are outside the \hull. Previous work has also reported the same for many other image datasets, both in pixel space and in feature space. This implies that any testing image has some novelty that is not captured in the training set. Overall, we can expect any new image that we receive to have some, at least minute, novelty to put it outside the \hull and $\mathcal{H}^{tr}_\phi$. 

Adversarial examples are usually created using a trained model, a model that is trained on a training set. It is not surprising then that adversarial methods actually move the images towards the convex hull of training set, i.e., the boundaries of their knowledge. Other adversarial methods such as poisoning attacks also work based on access to a training set, and as a result, they also come up with examples inside the \hull. Therefore, a testing image that falls inside or it is very close to the convex hull of training set in feature space can be considered suspicious if it is also close to a decision boundary in feature space.

% {\bf Other thoughts.} 
Based on current practices for developing adversarial examples, close proximity to decision boundaries in the feature space appears to be a good indicator to detect adversarial samples from genuine ones. We note, however, that it may be possible to design adversarial examples that remain far from the decision boundaries of feature space, and at the same time, maintain a distance from $\mathcal{H}^{tr}_\phi$.% This would require moving away from an image, crossing a decision boundary, maintaining a considerable margin from all the decision boundaries while remaining outside the convex hull of training set.

% \section{Introduction}

% This is where the content of your paper goes.
% \begin{itemize}
%   \item Limit the main text (not counting references and appendices) to 15 single-column PMLR-formatted pages, using this template.
%   \item Include, either in the main text or the appendices, all details, proofs and derivations required to substantiate the results.
%   \item Include {\em in the main text} enough details, %including proof details,
%     to convince the reviewers of the contribution, novelty and significance of the submissions.
%   \item Do not include author names (this is done automatically), and to the extent possible, avoid directly identifying the authors.
%     You should still include all relevant references, including your own, and any other relevant discussion, even if this might allow a reviewer to infer the author identities.
% \end{itemize}

% % Acknowledgments---Will not appear in anonymized version
% % \acks{We thank a bunch of people.}

% \bibliography{refs}

% \appendix

% \section{My Proof of Theorem 1}

% This is a boring technical proof.

% \section{My Proof of Theorem 2}

% This is a complete version of a proof sketched in the main text.

\end{document}